\pdfoutput=1
\documentclass[11pt]{article}

\usepackage[final]{acl}

\usepackage{times}
\usepackage{latexsym}

\usepackage[T1]{fontenc}
\usepackage[utf8]{inputenc}

\usepackage{microtype}

\usepackage{inconsolata}

\usepackage[]{footmisc}
\usepackage{amsmath}
\usepackage{cleveref}
\crefformat{section}{\S#2#1#3}
\crefformat{subsection}{\S#2#1#3}
\crefformat{subsubsection}{\S#2#1#3}
\crefrangeformat{section}{\S\S#3#1#4 to~#5#2#6}
\crefmultiformat{section}{\S\S#2#1#3}{ and~#2#1#3}{, #2#1#3}{ and~#2#1#3}

\Crefformat{figure}{#2Fig.~#1#3}
\Crefmultiformat{figure}{Figs.~#2#1#3}{ and~#2#1#3}{, #2#1#3}{ and~#2#1#3}
\Crefformat{table}{#2Tab.~#1#3}
\Crefmultiformat{table}{Tabs.~#2#1#3}{ and~#2#1#3}{, #2#1#3}{ and~#2#1#3}
\Crefformat{appendix}{#2Appx.~\S#1#3}
\crefformat{algorithm}{Alg.~#2#1#3}
\usepackage{graphicx}
\usepackage{caption}
\usepackage{subcaption}
\usepackage{booktabs}
\usepackage{multirow}
\usepackage{xspace}
\usepackage{enumitem}
\graphicspath{{imgs}}
\usepackage{algorithm}
\usepackage{algpseudocode}
\usepackage[]{mdframed}

\usepackage{amssymb}
\usepackage{pifont}
\usepackage{longtable}

\usepackage{xcolor}
\newcommand\crule[3][black]{\textcolor{#1}{\rule{#2}{#3}}}
\definecolor{c_0_shot}{HTML}{FDB462}
\definecolor{c_random}{HTML}{FB8072}
\definecolor{c_visual_clip}{HTML}{FFAEBC}
\definecolor{c_textual_clip}{HTML}{BEBADA}
\definecolor{c_bert}{HTML}{7570B3}
\definecolor{c_bertscore}{HTML}{8DD3C7}
\definecolor{c_MMICES}{HTML}{80B1D3}
\definecolor{c_ALBEF}{HTML}{868B8E}

\usepackage{url}

\usepackage{breakurl}
\usepackage[breaklinks]{hyperref}

\title{From Introspection to Best Practices: Principled Analysis of Demonstrations in Multimodal In-Context Learning}

\author{Nan Xu\textsuperscript{\protect\includegraphics[width=0.4cm]{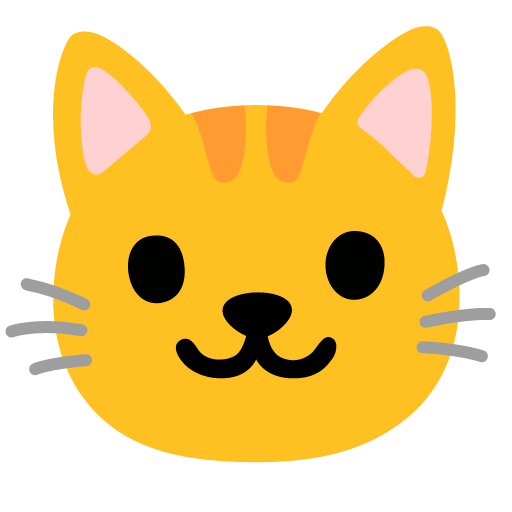}} \,Fei Wang\textsuperscript{\protect\includegraphics[width=0.4cm]{imgs/cat-face_1f431.png}}\, Sheng Zhang\textsuperscript{\protect\includegraphics[width=0.4cm]{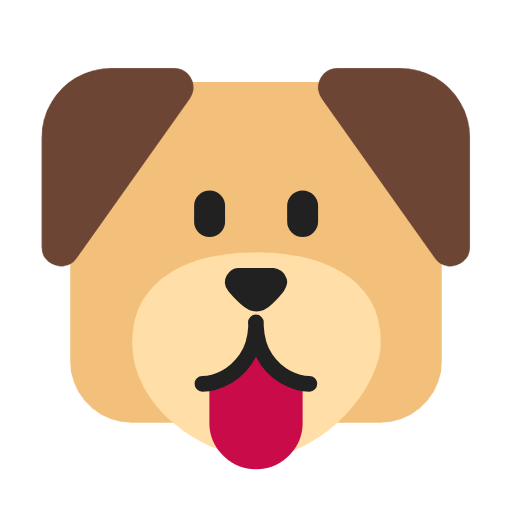}} \,Hoifung Poon\textsuperscript{\protect\includegraphics[width=0.4cm]{imgs/dog-face_1f436.png}}\,Muhao Chen\textsuperscript{\protect\includegraphics[width=0.4cm]{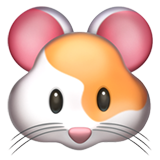}}\\
\textsuperscript{\protect\includegraphics[width=0.4cm]{imgs/cat-face_1f431.png}}University of Southern California \,\textsuperscript{\protect\includegraphics[width=0.4cm]{imgs/dog-face_1f436.png}}Microsoft Research\, \textsuperscript{\protect\includegraphics[width=0.4cm]{imgs/hamster_1f439.png}}University of California, Davis\\
\texttt{\{nanx,fwang598\}@usc.edu\, \{shezhan,hoifung\}@microsoft.com\, muhchen@ucdavis.edu
}}

\begin{document}
\maketitle
\begin{abstract}
Motivated by in-context learning (ICL) capabilities of Large Language Models (LLMs), multimodal LLMs with additional visual modality exhibit similar ICL abilities when multiple image-text pairs are provided as demonstrations. However, relatively less work has been done to investigate the principles behind how and why multimodal ICL works. We conduct a systematic and principled evaluation of multimodal ICL for models of different scales on a broad spectrum of new yet critical tasks. Through perturbations over different modality information, we show that modalities matter differently  across tasks in multimodal ICL. Guided by task-specific modality impact, we recommend  modality-driven demonstration strategies to boost ICL performance. 
% We also identify that demonstration selection is closely related to the models' ability to capture task inductive biases from multimodal ICL. 
We also find that models may follow inductive biases from multimodal ICL even if they are \emph{rarely seen} in or \emph{contradict semantic priors} from pretraining data.
Our principled analysis provides a comprehensive way of understanding the role of demonstrations in multimodal in-context learning, and sheds light on effectively improving multimodal ICL on a wide range of tasks.~\footnote{Codes and datasets are available at~\url{https://github.com/luka-group/MultimodalICLBestPractice}.}
% even if those tasks are not seen in or even contradict pretraining data.
\end{abstract}

\section{Introduction}
\begin{figure}[t!]
     \centering
         \begin{subfigure}[b]{\linewidth}
         \centering
         \includegraphics[width=\linewidth]{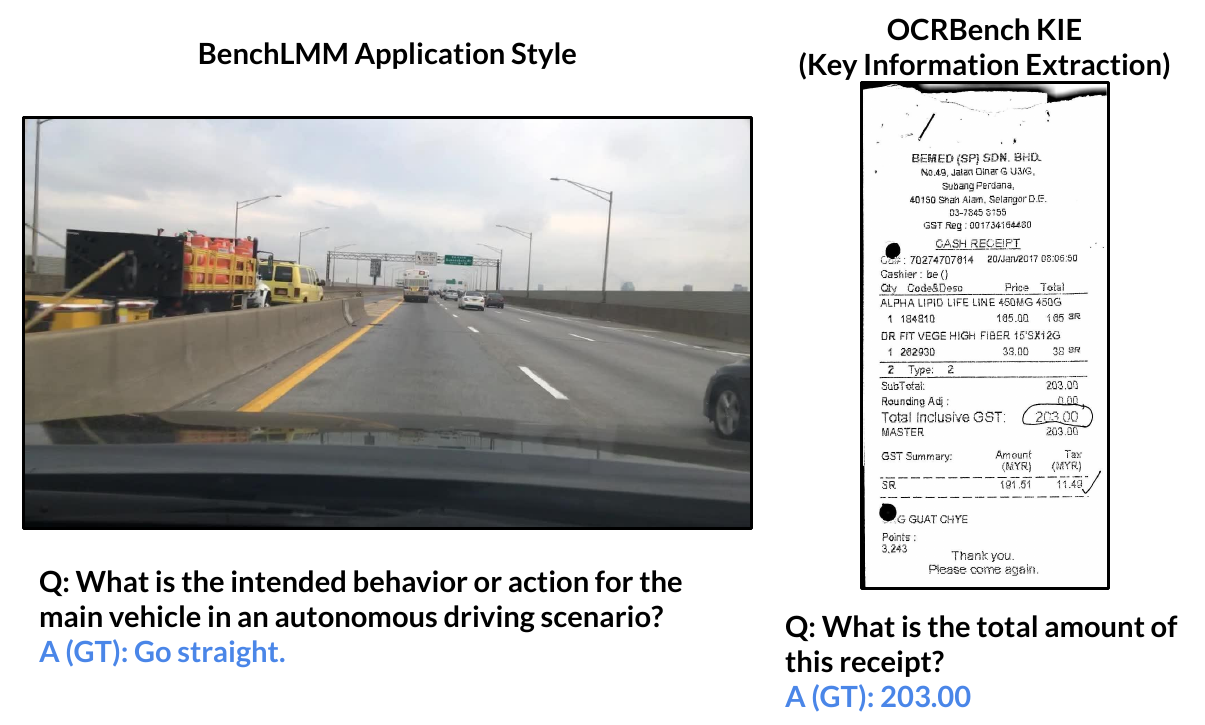}\caption{Questions and ground-truth answers from two of the investigated benchmarks: cross-style (left) and text-rich understanding (right).}
         \label{fig:case}
     \end{subfigure}
         \begin{subfigure}[b]{\linewidth}
         \centering
         \includegraphics[width=\linewidth]{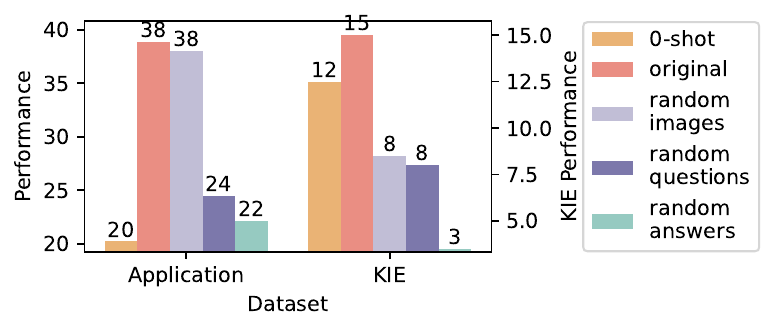}\caption{ICL performance against visual and textual perturbations.}
         \label{fig:4_demo_pereturbation}
     \end{subfigure}
              \begin{subfigure}[b]{\linewidth}
         \centering
         \includegraphics[width=\linewidth]{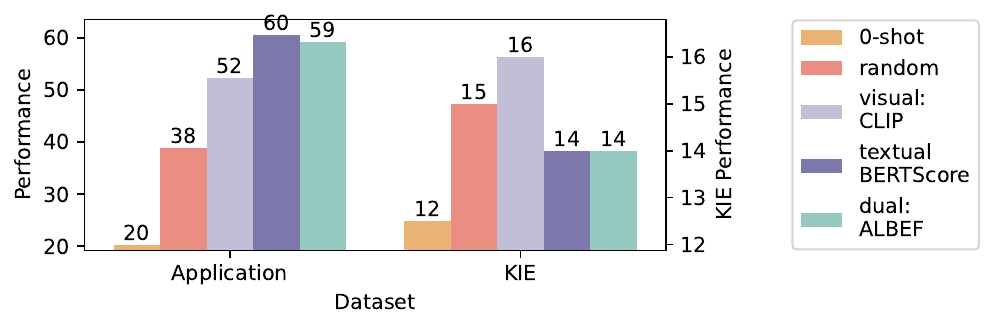}\caption{ICL performance given demonstrations selected by different modality-driven strategies.}
         \label{fig:4_demo_selection}
     \end{subfigure}
        \caption{(b) Modality matters differently on ICL across tasks: visual information matters little on Application but a lot on KIE, textual answers are more important to ICL on KIE than that on Application. (c) Demonstrations selected by text-driven strategy BERTScore benefit more on Application, while those selected by visual similarity (CLIP) bring higher accuracy on KIE.}
        \label{fig:demo}
        % \vspace{-1em}
\end{figure}

Motivated by in-context learning (ICL) capabilities of Large Language Models (LLMs) for NLP tasks~\citep{brown2020language,garg2022can,akyurek2022learning}, multimodal LLMs with additional visual modality are also exhibited with similar ICL abilities when multiple image-text pairs are provided as demonstrations~\citep{alayrac2022flamingo,bai2023qwen,sun2023generative,mckinzie2024mm1}. 
In recent studies, the Retrieval-based In-Context Example Selection
(RICES,~\citet{yang2022empirical}) approach, which retrieves similar images in
the support set by comparing their visual features with testing images, has become a default approach to selecting demonstrations for multimodal ICL~\citep{alayrac2022flamingo,sun2023generative,yang2024exploring}.
\begin{figure*}[t!]
     \centering
     \includegraphics[width=\textwidth]{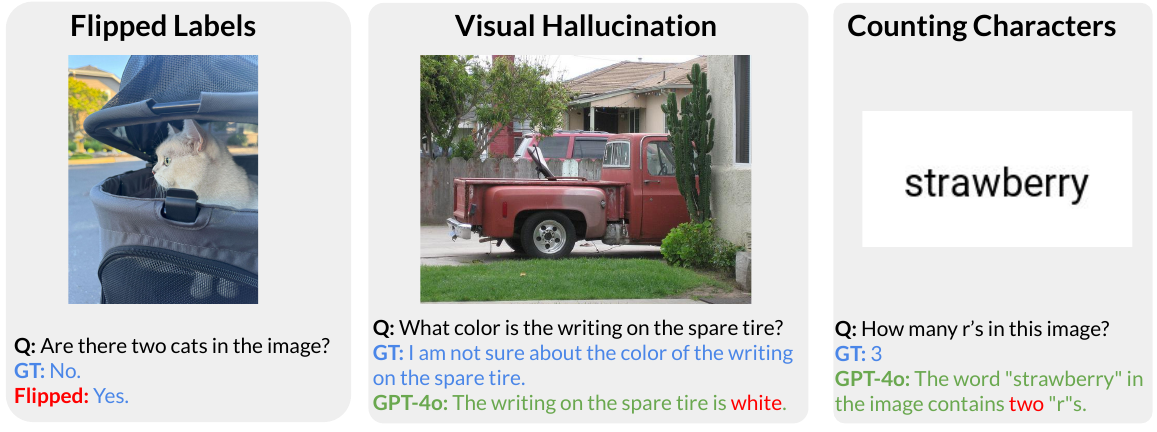}
        \caption{Benchmarks with inductive bias contracting the semantic priors (left) or rarely seen in pretraining data (middle and right). We list ground-truth (GT), zero-shot responses from GPT-4o and provide ICL analysis in~\Cref{sec:inductive}.}
        \label{fig:inductive_bias}
        % \vspace{-1em}
\end{figure*}
However, relatively less work has been done to investigate the principles behind how and why multimodal ICL works, nor has there been enough
justification for the necessity of selecting demonstrations according to visual modality and analyzing its advantages over other modalities. 
\citet{yang2024exploring} only explored better in-context configurations for image captioning, while \citet{chen2023understanding} argued that multimodal ICL is predominantly driven by the textual information in the demonstrations. However, their observations are limited to image captioning~\citep{young2014image,chen2015microsoft} and general-purpose visual question answering tasks~\citep{goyal2017making,gurari2018vizwiz,marino2019ok,sidorov2020textcaps}, which leaves a comprehensive exploration of the strengths of ICL and its limitations~\citep{zong2024vl} largely open for multimodal LLMs.

In this paper, we conduct a systematic and principled evaluation of multimodal ICL for models of different scales (ranging from OpenFlamingo 4B,~\citet{awadalla2023openflamingo} to IDEFICS1 80B,~\citet{laurencon2023obelics}) on a broad spectrum of new yet critical tasks as shown in~\Cref{fig:case}. These tasks require different types of capabilities, including hallucination mitigation~\citep{wang2023llm}, text-rich image understanding~\citep{liu2023hidden,li2024seed}, medical information comprehension~\citep{he2020pathvqa,pacheco2020pad,liu2021slake}, and cross-style transfer~\citep{cai2023benchlmm}, etc.

With diverse ICL capabilities examination, we show that the \textbf{dependency of performance gain from ICL on demonstration modalities differs among tasks} (\Cref{sec:modalities}). 
% As demonstrated in~\Cref{fig:4_demo_pereturbation}, 
For example, perturbing visual information in demonstrations (e.g., removing or replacing with random, noised or permuted images) does not cause significant performance drop on ICL for Application task (\Cref{fig:4_demo_pereturbation} left), while resulting in decreased accuracy than that provided by correct demonstrations on tasks such as key information extraction (KIE) from text-rich images (\Cref{fig:4_demo_pereturbation} right). 
% Sometimes it even leads to much worse performance than the zero-shot inference. 
On the other hand, textual perturbations (e.g., replacing the question/answer with random or one from other candidates in the same demonstration set) hurt ICL performance to different extents across tasks. 
% For instance,  perturbations on either questions or answers lead to greatly reduced accuracy on some tasks, while perturbations on answers results in extremely bad performance on others. 
These observations strongly suggest the necessity of understanding modality impact on ICL prior to collecting demonstrations  for specific tasks.

We conduct further investigation on how to select effective demonstrations to boost multimodal ICL performance (\Cref{sec:selection}). Based on empirical experiments, we recommend the following practices to elicit better ICL performance from multimodal models: \textbf{1) Utilizing demonstrations selected by visual similarity} (e.g., vision encoder of CLIP) for tasks observed with vital impact from visual modality on ICL performance, e.g., KIE task shown on the right of~\Cref{fig:4_demo_selection}. \textbf{2) Selecting demonstrations with high textual similarity} (e.g., text encoder of CLIP~\citep{radford2021learning} or BERTScore~\citep{zhang2019bertscore}) for tasks if the textual modality plays an important role in ICL performance, e.g., Application task shown on the left of~\Cref{fig:4_demo_selection}. \textbf{3) Choosing demonstrations with both visual and textual similarity considered} (e.g., ALBEF~\citep{li2021align} with a multimodal encoder that explicitly models interactions between image and text features) if dual modalities matter similarly to ICL performance. 
% As shown in~\Cref{fig:4_demo_selection}, we identify that providing demonstrations selected by textual similarity (e.g., text encoder of CLIP~\citep{radford2021learning} or BERTScore~\citep{zhang2019bertscore}) benefits ICL performance consistently across models and tasks. This is consistent with literature~\citep{chen2023understanding} and our prior observation that the textual modality plays an important role in ICL performance. For tasks observed with vital impact from visual modality on ICL performance, demonstrations selected by visual similarity (e.g., vision encoder of CLIP) elicit drastically improved ICL performance. Moreover, demonstration selection strategies that consider both visual and textual modalities, such as ALBEF~\citep{li2021align} with a multimodal encoder that explicitly models interactions between image and text features, present trade-off performance regardless of various modality importance to specific tasks.

Lastly, we illustrate that models may have the capability to capture task inductive biases from multimodal ICL (\Cref{sec:inductive}). 
We investigate two kinds of inductive bias: \emph{1) contradicting semantic priors}, and \emph{2) rarely seen in pretraining}. 
By deliberately flipping annotations of demonstrations to override strong semantic priors learned during pretraining (~\Cref{fig:inductive_bias} left), small-scale models fail to comprehend or follow practices provided by randomly sampled demonstrations, while they learn to follow inductive biases given demonstrations selected according to textual similarities, an emergent ability unlocked by scaling studied in literature~\citep{wei2022chain,zhou2022least}. Although multimodal LLMs are typically pretrained on true positives only but rarely on hallucination-inducing scenarios with unanswerable questions (~\Cref{fig:inductive_bias} middle, ~\citet{cha2024visually}), multimodal ICL greatly reduces hallucination over zero-shot inference. We also observe effectiveness of multimodal ICL in addressing the failure mode~\citep{ball2024can,shin2024large,yehudai2024can} of models on tasks humans find trivial (e.g., counting r's in the image plotting ``strawberry'' as shown in ~\Cref{fig:inductive_bias} right).
% we flip annotations of demonstrations to override semantic priors learned during pretraining (e.g., ``Yes'' to admit hallucinated objects in images and ``No'' to deny the presence  actually existing objects in images). Small-scale models fail to comprehend or follow practices against prior knowledge provided by randomly sampled demonstrations. Surprisingly, models learn to follow inductive biases given demonstrations selected according to textual similarities, an emergent ability unlocked by scaling studied in literature~\citep{wei2022chain,zhou2022least}. 
% This is reasonable as flipped annotations mainly convey inductive biases through texts. 
Such \textbf{capability to capture inductive bias of demonstrations without scaling up} models is more attractive than using semantic priors, since the model would be able to perform a wide range of tasks without further tuning, even if
those tasks are not seen in or even contradict pretraining data.

In summary, our principled analysis provides a comprehensive way of understanding the role of demonstrations in multimodal ICL. We empirically show that (1) modalities matter differently in multimodal ICL across tasks (\Cref{sec:modalities}), (2) demonstration strategies considering modality impact are able to boost ICL performance (\Cref{sec:selection}), (3) models are capable of capturing task inductive biases from multimodal ICL (\Cref{sec:inductive}). Overall, our work aims to shed light on effectively improving multimodal ICL on a wide range of tasks even if
those tasks are not seen in or even contradict pretraining data.

\section{Related Work}
\paragraph{Textual ICL}
LLMs have been recognized as strong few-shot learners since their emergence~\citep{brown2020language}.
With ICL, LLMs are empowered to generalize to a wide range of tasks at inference even if those tasks are not seen in pretraining data~\citep{garg2022can,akyurek2022learning}. 
% However, the performance of ICL is critically sensitive to the choices of demonstrations~\citep{rubin-etal-2022-learning,wang2023large,gupta-etal-2023-coverage}, the order~\citep{lu-etal-2022-fantastically,wu-etal-2023-self} and format of prompts~\citep{zhao2021calibrate,min2021noisy}. 
To understand why ICL works, 
% \citet{xie2021explanation} explained from the theoretical perspective that Transformers acquire the ICL ability when they are trained to infer latent concepts during pretraining. 
\citet{min-etal-2022-rethinking} empirically showed that the performance gain of ICL over zero-shot inference is mainly driven by the label space, distribution of input text, output labels, and overall format of the sequence, while the %ground-truth labels in demonstrations matter little
represented mapping from inputs to the outputs in demonstrations matters little. However, some recent work~\citep{zhou2022least,wei2023larger} suggested that when scaling up to some extent, larger models 
% (e.g. PaLM-540B~\citep{chowdhery2023palm} and Codex~\citep{chen2021evaluating}) 
can actually learn input-output mappings, which allows them to perform a variety of challenging tasks even if they contradict pretraining data.

Considering the additional visual information in multimodal ICL, we study the importance of different modalities and guide demonstration selection for better ICL performance accordingly.
\paragraph{Multimodal ICL}
After pretraining on interleaved image-text data or fine-tuning on multi-turn conversations, multimodal LLMs have exhibited ICL abilities in tasks such as image captioning and general-purpose visual question answering~\citep{alayrac2022flamingo,bai2023qwen,sun2023generative,mckinzie2024mm1}. Considering these studies may not sufficiently reveal strengths and weaknesses of ICL, \citet{zong2024vl} recently introduced VL-ICL Bench which encompasses a broad spectrum of tasks for multimodal ICL evaluation. However, there is not much work that conducts principled analysis on emergent ICL capabilities and provides insightful suggestions for future ICL practices. \citet{yang2024exploring} only explored better in-context configurations for image captioning. \citet{qin2024factors} recognized three factors--demonstration retrieval, ordering and instructions, that contribute to multimodal ICL performance, without providing task-specific  suggestions for demonstration selection to boost multimodal ICL capabilities.

One work that is closely connected to ours is~\citet{chen2023understanding}. \citet{chen2023understanding} argued that multimodal ICL is predominantly driven
by the textual information in the demonstrations and proposed Mixed Modality In-Context
Example Selection (MMICES), which first pre-filters samples
based on visual feature similarity and then selects most similar ones based on textual similarity. However, their observations are limited to image captioning and general-purpose visual question answering tasks, which leaves a comprehensive exploration for the strengths of ICL and its limitations~\citep{zong2024vl} largely open for multimodal LLMs. We conduct more comprehensive study on the impact of modality on ICL and find that modalities matter differently across tasks. Furthermore, we investigate how models of different scales capture task inductive biases from multimodal ICL.
\section{Experimental Setup}

In this section, we describe the experimental setup used in our analysis (\Cref{sec:modalities}-\Cref{sec:inductive}).
We list evaluation benchmarks and corresponding metrics in~\Cref{tab:dataset_statistics}, as well as studied model information in~\Cref{tab:model_details}. 
\paragraph{Evaluation Benchmarks}

After pretraining multimodal LLMs on interleaved image-text data or fine-tuning on multi-turn conversations, existing work~\citep{alayrac2022flamingo,bai2023qwen,sun2023generative,mckinzie2024mm1} mainly focuses on evaluating their ICL abilities on image captioning such as COCO~\citep{chen2015microsoft} and Flickr30K~\citep{young2014image}, as well as general-purpose visual question answering tasks such as OKVQA~\citep{marino2019ok}, VQAv2~\citep{goyal2017making}, TextVQA~\citep{sidorov2020textcaps} and VizWiz~\citep{gurari2018vizwiz}. Besides these classic vision-language tasks, we also consider one recently released benchmark, namely VL-ICL Bench~\citep{zong2024vl}, which encompasses a broad spectrum of challenging new tasks to investigate strengths and limitations of ICL capabilities.

% Although emergent in-context learning capabilities have been observed when evaluating in prior general-purpose benchmarks, 
Benefits of utilizing demonstrations as contexts for more critical and practical applications, though imperfect zero-shot performance is observed from state-of-the-art models, are not yet explored. Therefore, we further study ICL capabilities of multimodal LLMs on the following tasks. \emph{1) Math Reasoning}: MATH-Vision~\citep{wang2024measuring} is a large math reasoning benchmark that collects questions from real math competitions and tests the general visual perception and mathematical reasoning abilities; \emph{2) Hallucination}: AMBER~\cite{wang2023llm} provides a discriminative way to evaluate various types
of hallucination including existence,
attribute and relation; \emph{3) Text-rich Tasks}: both OCRBench~\citep{liu2023hidden} and SEED-Bench-2-Plus~\citep{li2024seed} assess text-rich visual comprehension of models, while the former focus on Optical Character Recognition (OCR) capabilities and the latter covers text-rich scenarios in the real world such as Charts, Maps, and Webs; \emph{4) Medical Tasks}: three datasets consider different medical modalities, i.e., Path-VQA~\citep{he2020pathvqa} for pathology, Slake-VQA~\citep{liu2021slake} for radiology and PAD-UFES-20~\citep{pacheco2020pad} for skin lesion images. \emph{5) Multi-image Tasks}: Seed-Bench-2~\cite{li2024seed} evaluates the ability to comprehend multimodal inputs containing multiple images. \emph{6) Cross-style Transfer}: BenchLMM~\citep{cai2023benchlmm} assesses the robustness of models against three different styles including artistic image, imaging sensor, and application styles.
\begin{figure*}[t!]
     \centering
    \begin{subfigure}[b]{\textwidth}
         \centering
         \includegraphics[width=\linewidth]{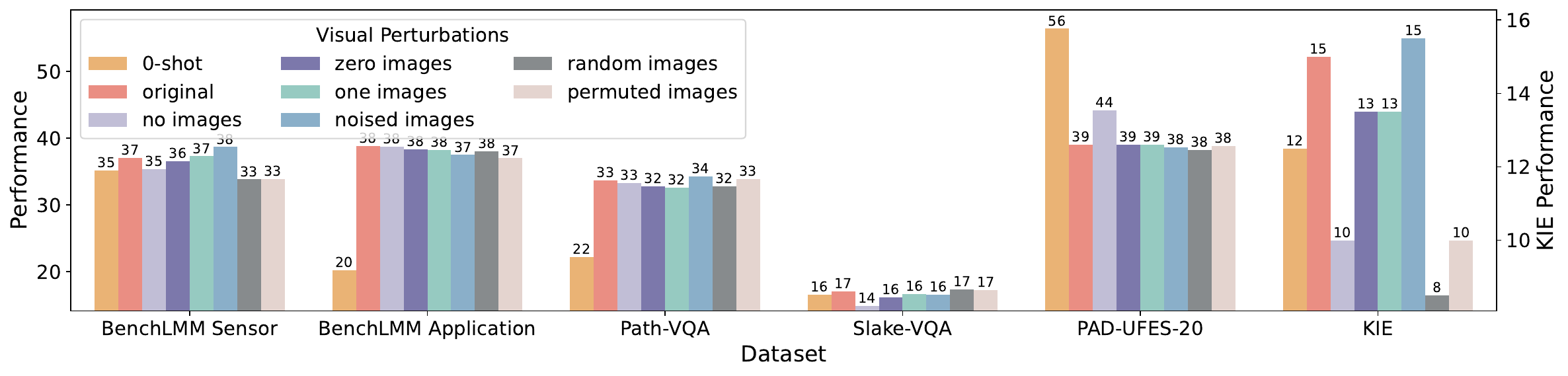}\caption{Visual Modality}
         \label{fig:visual_idefics-80b}
     \end{subfigure}
         \begin{subfigure}[b]{\textwidth}
         \centering
         \includegraphics[width=\linewidth]{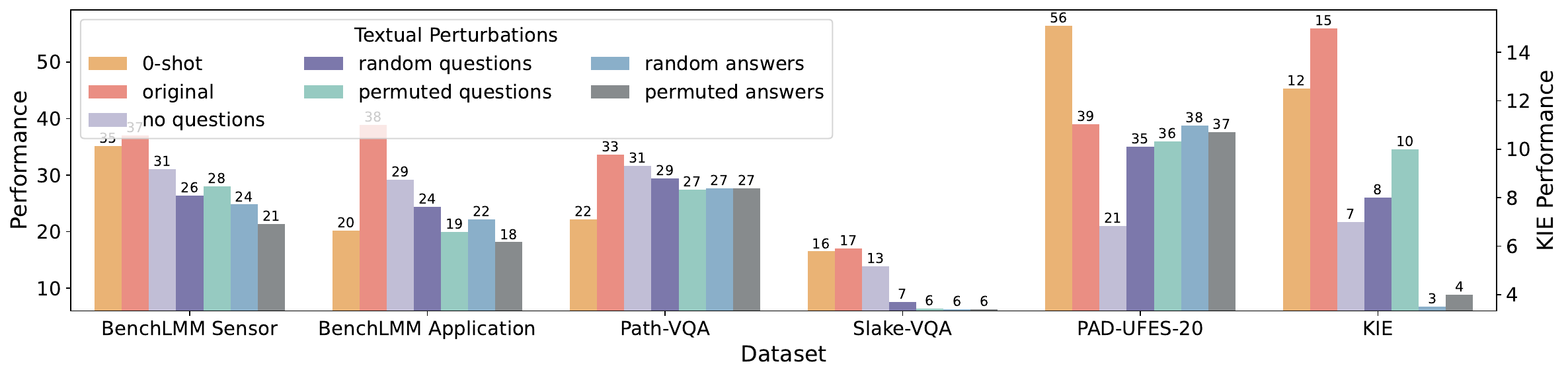}\caption{Textual Modality}
         \label{fig:textual_idefics-80b}
     \end{subfigure}
        \caption{Multimodal ICL performance of IDEFICS1-80b reacts differently across tasks of different difficulty levels against perturbations on visual (top) and textual (bottom) information. For  easy (i.e., \emph{BenchLMM Sensor} and \emph{Application}) and two moderate (i.e., \emph{Path-VQA} and \emph{Slake-VQA}) tasks, performance after various visual perturbations is very close to that given original correct demonstrations, while drops obviously when either textual question or answer is perturbed. For the moderate \emph{PAD-UFES-20}, neither of the two modalities matters too much. For the hard \emph{KIE}, we observe degraded performance when the image is removed or replaced. Similar observations from other 5 models can be found from~\Cref{fig:ablation_selection_openflamingo2_4B} to~\Cref{fig:ablation_selection_Emu1}.}
        \label{fig:ablation_idefics-80b}
        % \vspace{-1em}
\end{figure*}

\paragraph{Multimodal LLMs}

We evaluate pretrained multimodal LLMs without further instruction tuning, so that factors, such as seeing similar data or acquiring tested capabilities from the instruction dataset rather than through ICL, could be fairly reduced. Specifically, we consider the following pretrained models that scale from 4B to 80B and have previously demonstrated ICL abilities through limited analysis: OpenFlamingo~\citep{awadalla2023openflamingo} of two sizes (4B and 9B), IDEFICS of two scales from different versions (9B and 80B from the 1st version~\citep{laurencon2023obelics} and 8B from the 2nd version~\cite{laurençon2024matters}), together with the 14B Emu1~\citep{sun2023generative}.

Moreover, we evaluate the proprietary model, GPT-4o~\footnote{We use the version gpt-4o-2024-05-13 in this paper.}~\citep{openai2024gpt4o}, to exhibit challenge levels of evaluated tasks on the one hand, and compare ICL capabilities between pretrained and instruction-tuned models on the other hand.

% We provide detailed introduction to \textbf{Evaluation Metrics} and \textbf{Implementation Details} in~\Cref{sec:experimental_setup_more}.
\paragraph{Evaluation Metrics}
For image captioning, we report CIDEr~\citep{vedantam2015cider} scores. For general-purpose VQA tasks, we adopt the common VQA evaluation metric~\citep{antol2015vqa}, where $10$ annotations are provided and the model prediction is deemed $100\%$ accurate if at least three annotators provide that exact answer. To evaluate performance on two medical VQA task-slake-VQA and Path-VQA, we use the token-level F1 score following~\citet{tu2024towards}. We follow the evaluation practices in BenchLMM where ChatGPT is employed to gauge the proximity of answers predicted by the LMMs to ground-truth answers.
For remaining datasets, we utilize their original evaluation strategy--soft string matching, to eliminate the impact of answer formats. 
\paragraph{Implementation Details} We \emph{prompt} multimodal LLMs with an instruction ``Describe the image:'' for caption generation, while employing open-ended answer generation for other tasks with a prompt in the form of ``Question: the \texttt{<question>} Answer:'', without any constraint on model's output space.~\footnote{For short answer generation, we modify the prompt slightly to ``Question: \texttt{<question>} Short answer:''.} We adopt the default \emph{decoding} strategy and configurations (e.g., beam search with 5 as the number of beams for Emu1) suggested by each model vendor respectively. In contrast to the zero-shot setting, we consider 4- and 8-shot for in-context learning analysis~\footnote{Considering limited amounts of images per example used for pretraining, we evaluate 1- and 2-shot performance on tasks from SEED-Bench-2 where each example contains at least 8 images.}, where the demonstrations are \emph{randomly sampled} from candidates for each testing example unless otherwise stated.~\footnote{For each testing example, the demonstrations are randomly sampled from the train set while shared among all studied models.}
%\section{Tasks Matter Differently from Multimodal ICL}
\section{Modalities Matter Differently in Multimodal ICL}\label{sec:modalities}

As shown in~\Cref{fig:prior_icl} and~\Cref{fig:prior_icl_gpt}, pretrained models and GPT-4o generally achieve better performance given demonstrations as context in existing ICL tasks. As demonstrated in~\Cref{fig:new_icl}, on more %critical and practical
complex and reasoning-focused tasks, pretrained models %still benefit
generally benefit more from demonstrations while  performance of GPT-4o is barely influenced. 

In this section, we examine which modality of the demonstrations %leads to good performance of
takes more effect in multimodal in-context learning. 
For a comprehensive evaluation, we focus on three tasks of different difficulty levels: \emph{easy} cross-style tasks (i.e., BenchLMM Sensor and Application in~\Cref{fig:new_icl}), \emph{moderate} medical tasks (i.e., Path-VQA, Slake-VQA and PAD-UFES-20 in~\Cref{fig:new_icl}), and \emph{hard} text-rich key information extraction task (i.e., KIE from OCRBench in~\Cref{fig:ocr}). We visualize $4$-shot performance of \emph{IDEFICS-80b} within this section while leaving results of other models in Appendix (from~\Cref{fig:ablation_selection_openflamingo2_4B} to~\Cref{fig:ablation_selection_Emu1}.).
The dependency of performance gain from ICL on demonstration modalities differs among tasks.
% \nan{add a concise conclusion here}

% \footnotetext{In zero-shot setting, GPT-4o achieves extremely poor performance on general-purpose VQA datasets such as OKVQA, VQAv2, TextVQA and VizWiz. We find that GPT-4o tends to provide long answers even after we give the instruction ``Always provide short answers.'' as the system message. This results in low scores when comparing against short annotations.}

% image: two medical datasets minor; one medical no; two style yes; kie yes
% text: two medical yes; one medical no; style yes; kie yes.

\subsection{Impact of Visual Modality}\label{sec:visual_impact}
In recent studies, the Retrieval-based In-Context Example Selection
(RICES~\citep{yang2022empirical}) approach, which retrieves similar images in
the support set by comparing their visual features with testing images, has become a default approach to select demonstrations for multimodal in-context learning~\citep{alayrac2022flamingo,sun2023generative,yang2024exploring}. However, the necessity of selecting demonstrations according to visual modality and its advantages over others is not yet explored.

By fixing the textual modality (i.e., question and answer pairs) of demonstrations, we experiment with demonstrations containing different %variations
perturbations of visual modality: \emph{1) no images} where only textual question and answer pairs are provided; \emph{2) zero/one images} that all zero (black)/255(white) pixel values are used instead; \emph{3) noised images} that apply Gaussian noises to the original images; \emph{4) random images} sampled from the train set; \emph{5) permuted images} reorganize the order of demonstration images to misalign visual/textual modalities. 
\paragraph{Results} 
We compare ICL performance of IDEFICS1-80B~\footnote{We show performance of IDEFICS1-80B in ~\Cref{fig:ablation_idefics-80b}  on all tasks except KIE, which is too challenging for IDEFICS1-80B to handle in both zero- and few-shot settings (at most 3 of 200 testing examples are answered correctly). Only IDEFICS2-8b can solve considerable amounts of cases (30 out of 200 in 4-shot setting), hence we perturb modality information on IDEFICS2-8b instead.} before and after visual perturbations in~\Cref{fig:ablation_idefics-80b} and other models from~\Cref{fig:visual_openflamingo2_4B} to~\Cref{fig:visual_Emu1}. For easy cross-style and moderate medical tasks, we find that perturbing visual information in demonstrations does not cause significant performance drop on ICL, which is consistent with observations from prior work~\citep{chen2023understanding}. However, for the hard KIE task, visual perturbations that remove or change content of images result in decreased accuracy than that provided with correct demonstrations, sometimes  much worse performance than the zero-shot inference. This indicates that visual information plays an important role in improving ICL performance over zero-shot one, which is reasonable since this dataset
requires extracting key-value pairs in the image~\citep{liu2023hidden}. Meanwhile, the performance after applying Gaussian noises to images is very close to performance with correct images, which implies that multimodal LLMs are agnostic to image noises and able to extract key visual information for question answering.

\subsection{Impact of Textual Modality}\label{sec:textual_impact}
Previous studies have identified excessive dependence of multimodal LLMs on the language model's linguistic priors~\citep{han2022visual,li2023evaluating}. Accordingly, the role of textual modality for multimodal ICL should be similarly important. Therefore, we keep the visual modality of demonstrations while performing the following %variations
perturbations upon textual question and answer pairs: \emph{1) no questions/answers} remove the question/answer component directly; \emph{2) random questions/answers} employ questions/answers sampled from the train set instead; \emph{3) permuted questions/answers} exchange question or answer component of demonstration examples while keeping the other two components unchanged.
\paragraph{Results}
In~\Cref{fig:textual_idefics-80b}, we visualize ICL performance in response to perturbations upon questions or answers of demonstrations independently. We find that textual perturbations hurt ICL performance to different extents. On tasks such as BenchLMM Sensor, Slake-VQA and KIE, perturbations on either questions or answers lead to greatly reduced accuracy even below zero-shot inference. By replacing correct answers from demonstrations with random ones or those misaligned with image-question pairs, we observe extremely bad performance on Slake-VQA and KIE. On other tasks, questions and answers are almost equally important to  ICL.

%\section{Why does In-context Learning Work?}
%\section{Modalities Matter Differently in Multimodal ICL}

\section{How to Select Effective Demonstrations for Multimodal ICL}\label{sec:selection}
\begin{figure*}[t!]
     \centering
     \includegraphics[width=\linewidth]{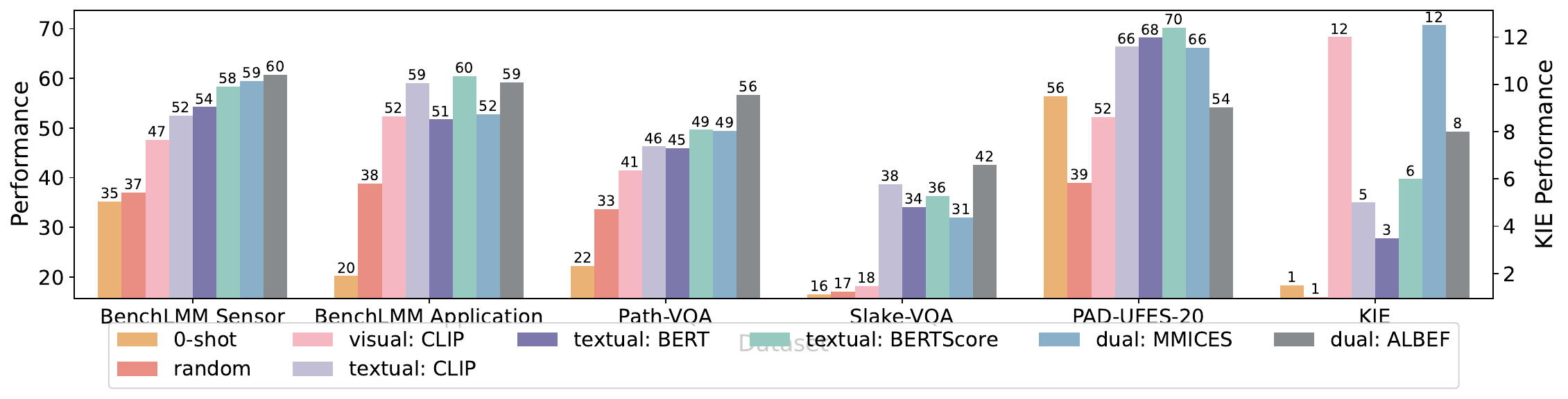}
        \caption{Influence of modality-driven demonstration selection strategies on ICL performance of IDEFICS1-80B. Text-driven demonstration (e.g., \crule[c_textual_clip]{.3cm}{.3cm} textual CLIP, \crule[c_bert]{.3cm}{.3cm} BERT, and \crule[c_bertscore]{.3cm}{.3cm} BERTScore) selection strategies always bring performance improvement over zero-shot inference and random strategy. Strategies considering visual modality (e.g., \crule[c_visual_clip]{.3cm}{.3cm} visual CLIP, \crule[c_MMICES]{.3cm}{.3cm} MMICES, and \crule[c_ALBEF]{.3cm}{.3cm} ALBEF) enhance performance significantly on KIE, where visual modality proves to be critical for ICL performance as illustrated in~\Cref{fig:visual_idefics-80b}. We visualize similar observations of other five models from~\Cref{fig:ablation_selection_openflamingo2_4B} to~\Cref{fig:ablation_selection_Emu1}.}
        \label{fig:selection_idefics-80b}
        % \vspace{-1em}
\end{figure*}
Motivated by variational roles of different modalities across different tasks, we further explore influence of modality-driven demonstration selection strategies on ICL performance in this section.

\paragraph{Vision-driven Demonstration Selection} To retrieve demonstrations containing images similar to those in testing examples, we follow prior studies~\citep{alayrac2022flamingo,sun2023generative,yang2024exploring} by adopting the RICES strategy~\citep{yang2022empirical}, which compares visual similarity according to features extracted from the pretrained visual encoder of CLIP~\citep{radford2021learning}.

\paragraph{Text-driven Demonstration Selection} For fair comparison with RICES, we employ the textual encoder of CLIP as well for selecting demonstrations with similar textual features to testing examples. We also adopt the BERTScore~\citep{zhang2019bertscore} metric~\footnote{We adopt the DeBERTa large model fine-tuned with MNLI task, which is accessible at \url{https://huggingface.co/microsoft/deberta-large-mnli}.}, which considers token-level similarity between candidate and reference sentences and shows a strong correlation with human judgements on multiple common benchmarks.
\paragraph{Dual-modality driven Demonstration Selection} We first consider Mixed Modality In-Context
Example Selection (MMICES) proposed by~\citet{chen2023understanding}, which first pre-filters 
$K$ samples ($K$=32) based on visual feature similarity and then selects the most similar ones based on textual similarity. To represent vision-language features, we utilize ALBEF~\citep{li2021align}, a multimodal encoder that explicitly models the interactions between image and text features and achieves state-of-the-art performance on image-text retrieval tasks. Since its multimodal encoder is built upon an image encoder (i.e., visual transformer ViT-B/16) and a text encoder (i.e., $\text{BERT}_{\text{base}}$), we also select demonstrations according to the embedding of the \texttt{[CLS]} token from $\text{BERT}_{\text{base}}$ as another textual-driven approach for contrast. For fair comparison, the vision-driven CLIP approach, the visual feature extractor of MMICES, and the visual encoder of ALBEF share the same visual transformer (i.e., ViT-B/16).

% We focus on demonstration selection in this paper.
Considering the sensitivity of LLMs to the ordering in the prompt~\citep{lu-etal-2022-fantastically,wu-etal-2023-self}, we follow prior work~\citep{alayrac2022flamingo,gupta-etal-2023-coverage} with demonstrations ordered by an increasing order of similarity, such that the most similar demonstration appears right before the testing example.

\paragraph{Results}
We illustrate influence of demonstration selection strategies on ICL performance in~\Cref{fig:selection_idefics-80b}. Providing demonstrations selected by textual similarity benefits ICL performance consistently across models and tasks. This is consistent with literature~\citep{chen2023understanding} and our observations in~\Cref{sec:textual_impact} that the textual modality plays an important role in ICL performance. In general, the larger text embedding model--BERTScore (124M parameters) leads to better ICL performance compared with smaller models like textual CLIP (63M parameters) and BERT (124M parameters).

As analyzed in~\Cref{sec:visual_impact}, visual information of demonstrations is of vital importance to ICL performance for the task KIE that requires key-value pair extraction from images. Accordingly, we witness drastically improved ICL performance when demonstrations containing more similar images to testing images are provided by visual CLIP. 

Strategies that consider dual modalities for demonstration selection (e.g., MMICES and ALBEF) are similarly more advantageous compared with text-driven methods on KIE. We also find that they achieve trade-off performance regardless of various modality importance to specific tasks. Meanwhile, ALBEF which explicitly models the interactions between image and text features obtains better ICL performance than MMICES, which is constrained by the vision-driven pre-filter process.

%\section{How to Select Effective Demonstrations for Multimodal ICL}

% \section{Models May Not Always Capture Task Inductive Biases from Multimodal ICL}\label{sec:inductive}

\section{Models May Capture Task Inductive Biases from Multimodal ICL}\label{sec:inductive}
\begin{figure*}[t!]
     \centering
     \includegraphics[width=\linewidth]{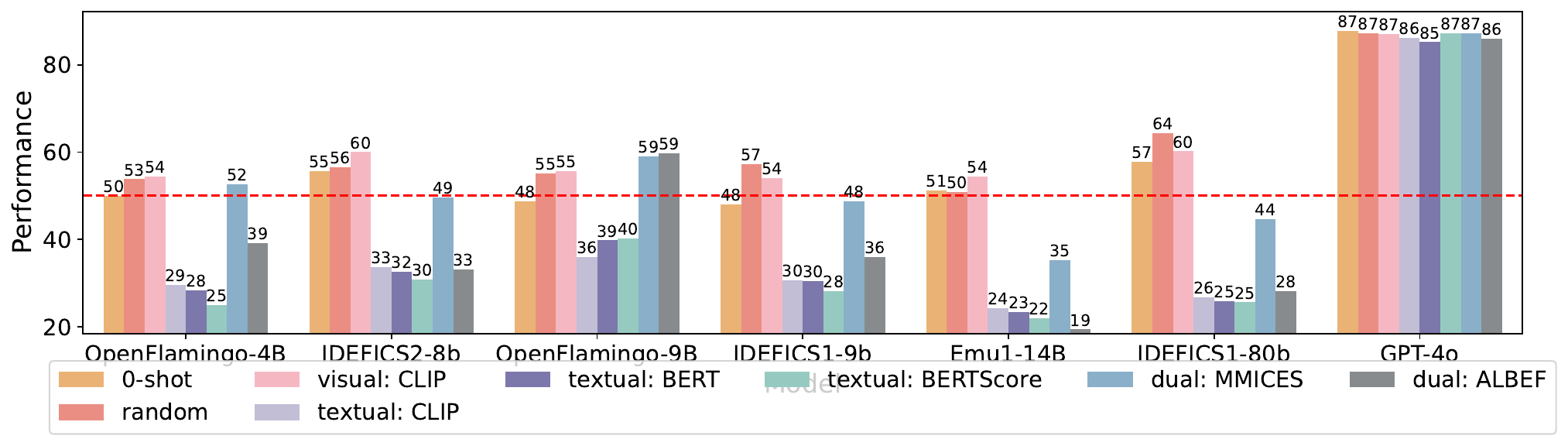}
        \caption{The ability to capture inductive biases that contradict semantic priors when presented with flipped in-context exemplar
annotations of AMBER Attribution emerges when demonstrations are selected according to \emph{textual} modality (i.e., \crule[c_textual_clip]{.3cm}{.3cm} textual CLIP, \crule[c_bert]{.3cm}{.3cm} BERT and \crule[c_bertscore]{.3cm}{.3cm} BERTScore). Ground truth annotations for testing examples are not flipped, so if a model learns to
follow flipped labels in demonstrations, its accuracy should be below $50\%$. Given random demonstrations \crule[c_random]{.3cm}{.3cm} or those selected considering visual modality \crule[c_visual_clip]{.3cm}{.3cm}, models cannot flip predictions to follow flipped annotations, while models can do so provided with demonstrations selected by text-driven strategies (performance decreases to well
below $50\%$). We show similar observations on Existence and Relation in~\Cref{fig:AMBER_flip}.}
        \label{fig:AMBER_attribute_flip}
        % \vspace{-1em}
\end{figure*}
Prior work on NLP tasks shows that small language models like GPT-J-6B~\citep{wang2021gpt}
% , PaLM-8B~\citep{chowdhery2023palm} and GPT3 curie-6.7B~\citep{gao2021framework} 
and PaLM-8B~\citep{chowdhery2023palm}
rely primarily on semantic priors from pretraining~\citep{min-etal-2022-rethinking}, while large models such as PaLM-540B
% , InstructGPT~\citep{ouyang2022training} and Codex~\citep{chen2021evaluating} 
and InstructGPT~\citep{ouyang2022training}
can capture and follow inductive biases from in-context exemplars, performing a wide range of tasks even if those tasks are not seen~\citep{garg2022can,akyurek2022learning} in or even contradict~\citep{wei2023larger} pretraining data. 
% even when they contradict strong semantic priors that larger models may hold~\citep{wei2023larger}. 
However, it is unknown whether capturing inductive biases is still an emergent ability
of model scale for multimodal ICL. 
We experiment with three benchmarks where the inductive bias contradicts semantic priors (~\Cref{fig:inductive_bias} left) or are rarely seen (~\Cref{fig:inductive_bias} middle \& right) in pretraining.

\begin{table*}[t!]
\centering
\resizebox{\textwidth}{!}{%
\begin{tabular}{@{}lccccccc@{}}
\toprule
\bf Setting            & \bf OpenFlamingo-4B & \bf IDEFICS2-8b   & \bf OpenFlamingo-9B & \bf IDEFICS1-9b   & \bf Emu1-14B      & \bf IDEFICS1-80b  & \bf GPT-4o        \\ \midrule
\multicolumn{8}{c}{\bf Visual Hallucination} \\
\bf 0-shot             & 12.0            & 0.6           & 17.4            & 0.6           & 0.4           & 2.0           & 28.2          \\%\midrule
\bf Random             & 76.2            & 37.4          & 73.2            & 61.4          & 37.6          & 59.6          & 33.4          \\%\midrule
\bf visual: CLIP       & 79.2            & 38.8          & 71.6            & 68.6          & 49.4          & 64.4          & 40.0          \\%\midrule
\bf textual: CLIP      & 79.2            & 49.0          & 70.4            & 66.0          & 53.2          & 62.0          & 41.2          \\
\bf textual: BERT      & \textbf{85.6}   & \textbf{63.2} & \textbf{75.0}   & \textbf{74.8} & \textbf{60.2} & 74.4          & \textbf{46.6} \\
\bf textual: BERTScore & 84.2            & 62.2          & 74.6            & 74.4          & 57.8          & \textbf{75.8} & 44.2          \\%\midrule
\bf dual: MMICES       & 78.8            & 43.6          & 70.4            & 69.4          & 49.8          & 64.6          & 39.8          \\
\bf dual: ALBEF        & 83.4            & 55.8          & 71              & 72.6          & 57.0          & 70.0          & 43.4          \\ \midrule\midrule
\multicolumn{8}{c}{\bf Counting Characters} \\
\textbf{0-shot}             & 19.6                     & 68.8                 & 9.4                      & 60.6                 & 24.6              & 36.4                  & 77.4            \\%\midrule
\textbf{Random}             & 70.2                     & 78.6                 & \textbf{60.4}            & 75.0                 & \textbf{79.4}     & 76.6                  & 91.4            \\%\midrule
\textbf{visual: CLIP}       & \textbf{70.6}            & \textbf{79.0}        & 57.4                     & \textbf{76.6}        & 78.8              & 76.2                  & 90.6            \\%\midrule
\textbf{textual: CLIP}      & 68.0                     & \textbf{79.0}        & 48.8                     & 75.6                 & \textbf{79.4}     & \textbf{78.0}         & 88.0            \\
\textbf{textual: BERT}      & 56.4                     & 77.4                 & 38.2                     & 75.4                 & 79.4              & 70.0                  & 85.8            \\
\textbf{textual: BERTScore} & 58.8                     & 75.4                 & 39.6                     & 76.0                 & 79.0              & 73.6                  & 88.0            \\%\midrule
\textbf{dual: MMICES}       & 56.8                     & 75.8                 & 41.4                     & 70.6                 & 77.8              & 69.8                  & \textbf{91.8}   \\
\textbf{dual: ALBEF}        & 66.2                     & 78.0                 & 40.4                     & 73.6                 & 77.8              & 74.0                  & 84.6            \\ \bottomrule
\end{tabular}%
}
\caption{The capability to capture inductive biases that are rarely seen in pretraining emerges in multimodal ICL. Multimodal LLMs are able to capture and follow inductive bias, hence reducing hallucination in responses to unanswerable questions (\textbf{top}) and more accurately counting characters within queried words (\textbf{bottom}).}
\label{tab:rarely_seen_inductive_bias}
        % \vspace{-1em}
\end{table*}
% We provide detailed introduction to utilized \textbf{benchmarks} (i.e., \emph{flipped labels}, \emph{visual hallucination} and \emph{counting characters}) in~\Cref{sec:inductive_benchmarks} and \textbf{implementation details} in~\Cref{sec:inductive_implementation_details}.
\subsection{Benchmarks}~\label{sec:inductive_benchmarks}
\paragraph{Flipped Labels} We flip original labels from the hallucination benchmark AMBER~\citep{wang2023llm}. The selected demonstration is labeled as ``Yes'' if the description in the question is WRONG according to the image, ``No'' otherwise.  
\paragraph{Visual Hallucination} We adopt the VQAv2-IDK benchmark~\citep{cha2024visually}. It contains hallucination-inducing scenarios, where providing definitive answers is challenging and responses such as ``I Don't Know'' are desired. Multimodal LLMs are typically pretrained on true positives only but rarely on such hallucination-inducing scenarios, hence striving to answer with hallucination. 
\paragraph{Counting Characters} Recent state-of-the-art LLMs are capable of performing complex reasoning~\citep{llama2024}, math problem-solving~\citep{qwen2_math}, code generation~\citep{team2024codegemma} and even challenging Mathematical Olympiad (IMO) tasks~\citep{deepmind_imo_2024}, but fail to handle problems that humans find trivial, e.g., counting the number of r's in the word ``strawberry'' or ``barrier''~\citep{ball2024can,shin2024large,yehudai2024can}.
% , which is called jagged intelligence~\citep{karpathy_tweet_2024_jagged}. 
Interestingly, we find it similarly challenging for multimodal LLMs to count the occurrence of characters when the word is displayed as an image (i.e., individual black word plotted on white background). 
% We draw an image containing black word in white background, and ask models to answer the occurrence of one letter. 
We investigate whether multimodal LLMs can discover and follow the inductive bias of character counting from demonstrations, which is probably rarely seen during model pretraining.

% We investigate both random and distinct modality-driven demonstration selection strategies to analyze the relation of capturing inductive biases from ICL to model scales and demonstration quality.
\subsection{Implementation Details}\label{sec:inductive_implementation_details}
We evaluate $500$ testing instances in the $8$-shot setting for all benchmarks, where demonstrations per testing instance are randomly sampled from $5,000$ candidates~\footnote{On VQAv2-IDK, we test on $500$ unanswerable questions with demonstrations selected from a set of $5,000$ answerable and $5,000$ unanswerable questions to ensure models are not guided to reject to answer no matter what question is presented.}. For \emph{Flipped Labels} where inductive bias conflicts with semantic priors, we compare $8$-shot performance with the random guess (i.e., $50\%$ accuracy for the yes/no questions). For the other two benchmarks where inductive bias from demonstrations is rarely seen during pretraining and expected to help models better perceive and address tasks, we compare few-shot with zero-shot results to learn whether inductive bias is captured. 
% \subsection{Results} 
\subsection{Induct Bias Contradicting Semantic Priors}We show the abilities of different models for capturing inductive biases from demonstrations in~\Cref{fig:AMBER_attribute_flip}. We flip annotations of demonstrations while keeping the ground-true answers of testing examples unflipped, hence the lower the accuracy, the stronger the capabilities of multimodal LLMs to capture inductive biases and further override semantic priors learned during pretraining. When provided with demonstrations randomly sampled or selected according to similarities of visual features (i.e., visual CLIP), all evaluated models fail to comprehend or follow practices against prior knowledge. This is consistent with existing studies showing that small language models ignore flipped labels presented in-context and thus rely primarily on semantic priors from pretraining~\citep{wei2023larger}. Surprisingly, all studied small-scale models tend to follow inductive biases from demonstrations with accuracy well below $50\%$ when we switch demonstrations to those selected according to textual similarities (e.g., textual CLIP, BERT, BERTScore). We suspect that flipped annotations mainly convey inductive biases through texts, which makes text-driven selection strategies effective in guiding the behavior of small models to override semantic priors.

Notably, GPT-4o always follows the strong semantic priors and provides factual responses even when the demonstration annotations are flipped, which is quite opposite to the emergent ability unlocked of model scale discovered in the literature~\citep{wei2022emergent,wei2023larger}. However, GPT-4o's failure to provide flipped answers following demonstrations does not indicate such a large model is unable to capture those inductive biases. We speculate that GPT-4o may be able to perceive provided biases that are against semantic priors, but reject to give non-factual responses due to its built-in safety mechanisms across modalities~\citep{openai2024gpt4o}.
\subsection{Inductive Bias Rarely Seen in Pretraining}
In~\Cref{tab:rarely_seen_inductive_bias}, we study whether models can capture inductive bias rarely seen in pretraining in multimodal ICL. In zero-shot setting, we observe quite poor performance consistently across studied models on two datasets.
% learn to recognize the unanswerability of questions from multimodal ICL, rather than striving to answer even if the response is hallucinated. 
Although current models rarely see unanswerable questions during pretraining, we observe emergent abilities to answer unanswerable questions without hallucination in multimodal ICL (top in~\Cref{tab:rarely_seen_inductive_bias}), especially with demonstrations selected by text-driven strategies. 
% \paragraph{Counting Characters} 
As shown at the bottom of ~\Cref{tab:rarely_seen_inductive_bias}, the accuracy of character counting improves greatly in ICL compared with zero-shot. Vision-driven demonstration selection strategy CLIP is more effective than others. Text-driven strategies do not achieve top performance, which is reasonable since key information to answer questions comes from letters in the image, rather than the text part.

\section{Conclusion}
We conduct a systematic and principled evaluation of multimodal ICL for models of different scales on a broad spectrum of new yet critical tasks. We find that modalities matter differently in multimodal ICL across tasks. Hence we utilize modality-driven demonstration strategies to boost ICL performance. Moreover, we find that demonstrations selected according to textual similarity help models capture inductive biases from multimodal ICL.
\clearpage
\section*{Limitations}
We conduct a systematic and principled evaluation of multimodal ICL for pretrained models of different scales on a broad spectrum of new yet critical tasks. One limitation of our study is lack of discussion over instruction-tuned models, which may present differently than pretrained ones.
\section*{Ethics Statement}
This paper presents comprehensive study of multimodal ICL on multiple existing benchmarks that have gone through ethical reviews in prior works. Therefore, we believe our work does not pose additional ethical issues.
\section*{Acknowledgements}
We appreciate the reviewers for their insightful comments and suggestions.

Muhao Chen was supported by the DARPA FoundSci Grant HR00112490370 and the NSF of the United States Grant ITE 2333736.
% \bibliographystyle{acl_natbib}
% \bibliography{anthology,custom}
\bibliography{custom}
\clearpage

\appendix
\section{Appendix}\label{sec:appendix}
% \subsection{Experimental Setup}~\label{sec:experimental_setup_more}

% \section{Models May Capture Task Inductive Biases from Multimodal ICL}\label{sec:inductive_more}

% To unveil the mystery behind the failure in counting characters from multimodal models, we randomly sample $500$ words with demonstrations sampled from $5,000$ instances.
\begin{table*}[t!]
\resizebox{\textwidth}{!}{%
\begin{tabular}{@{}llcccc@{}}
\toprule
Capabilities Tested                                                                                             & Dataset              & \#Train   & \#Test & Metric   & References \\ \midrule
\multirow{2}{*}{Captioning Image}                                                                               & COCO                 & 2,815,816 & 500    & CIDEr    &    \citet{chen2015microsoft}        \\& Flickr30K            & 29,000    & 500    & CIDEr    & \citet{young2014image}           \\\midrule
\multirow{4}{*}{\begin{tabular}[c]{@{}l@{}}General visual perception \\ and textual understanding\end{tabular}} & OKVQA                & 9,009     & 500    & Accuracy &   \citet{marino2019ok}         \\& VQAv2                & 443,757   & 500    & Accuracy &  \citet{goyal2017making}          \\& TextVQA              & 34,602    & 500    & Accuracy & \citet{sidorov2020textcaps}           \\& VizWiz               & 20,523    & 500    & Accuracy &   \citet{gurari2018vizwiz}         \\\midrule
In-context Learning                                                                                             & VL-ICL$^*$              & 9,960     & 1,120  & Accuracy &   \citet{zong2024vl}         \\\midrule\midrule
Mathematical Reasoning                                                                                          & MATH-Vision          & 2,540     & 500    & Accuracy &  \citet{wang2024measuring}          \\\midrule
\multirow{3}{*}{Hallucination}                                                                                  & AMBER Existence      & 8,763     & 500    & Accuracy &  \multirow{3}{*}{\citet{wang2023llm}}         \\& AMBER Attribute      & 7,124     & 500    & Accuracy &            \\& AMBER Relation       & 1,163     & 500    & Accuracy &            \\\midrule
\multirow{2}{*}{\begin{tabular}[c]{@{}l@{}}Text-rich Visual \\ Comprehension\end{tabular}}                                                                & OCRBench$^*$            & 53,991    & 900    & Accuracy &  \citet{liu2023hidden}          \\& SEED-Bench-2-Plus$^*$   & 1,174     & 1,103  & Accuracy &  \citet{li2024seed}          \\\midrule
\multirow{3}{*}{Medical}                                                                                        & Path-VQA             & 19,755    & 500    & Token F1 &    \citet{he2020pathvqa}       \\& Slake-VQA            & 9,835     & 500    & Token F1 & \citet{liu2021slake}           \\& PAD-UFES-20          & 994       & 500    & Accuracy &    \citet{pacheco2020pad}        \\\midrule
Multiple Images                                                                                                 & Seed-Bench-2         &   3,751        &   2,260     & Accuracy &    \citet{li2024seed}        \\\midrule
\multirow{3}{*}{Cross-style}                                                                                    & BenchLMM Artistic    &     100      &    400    & Accuracy &    \multirow{3}{*}{\citet{cai2023benchlmm}}         \\& BenchLMM Sensor      &300           & 400       & Accuracy &            \\& BenchLMM Application &  367         & 400       & Accuracy &            \\\bottomrule 
\end{tabular}}
\caption{Evaluation benchmark statistics. We adopt the default train and test split as the demonstration candidates and testing examples if the testing annotations are provided, otherwise the validation split is used instead. We randomly sample at most $500$ instances for testing. The three datasets marked by~$^*$ are composed of multiple subsets and we consider average performance for analysis, leaving detailed results in Appendix.}
\label{tab:dataset_statistics}
\end{table*}
\begin{table*}[t!]
\resizebox{\textwidth}{!}{%
\begin{tabular}{@{}llll@{}}\toprule
Multimodal LLMs                   & Visual Encoders                       & LLMs                                                                & \#Params \\\midrule
                                  & openai CLIP ViT-L/14                  & togethercomputer/RedPajama-INCITE-Base-3B-v1                        & 4B           \\
\multirow{-2}{*}{OpenFlamingo-4B} & \multicolumn{3}{l}{\url{https://huggingface.co/openflamingo/OpenFlamingo-4B-vitl-rpj3b}}                                         \\\midrule
                                  & openai CLIP ViT-L/14                  & anas-awadalla/mpt-7b & 9B           \\
\multirow{-2}{*}{OpenFlamingo-9B} & \multicolumn{3}{l}{\url{https://huggingface.co/openflamingo/OpenFlamingo-9B-vitl-mpt7b}}                                         \\\midrule
                                  & laion/CLIP-ViT-H-14-laion2B-s32B-b79K & huggyllama/llama-7b                                                 & 9B           \\
\multirow{-2}{*}{IDEFICS1-9B}     & \multicolumn{3}{l}{\url{https://huggingface.co/HuggingFaceM4/idefics-9b}}                                                        \\\midrule
                                  & laion/CLIP-ViT-H-14-laion2B-s32B-b79K & huggyllama/llama-65b                                                & 80B          \\
\multirow{-2}{*}{IDEFICS1-80B}    & \multicolumn{3}{l}{\url{https://huggingface.co/huggyllama/llama-65b}}                                                            \\\midrule
IDEFICS2-8B                       & google/siglip-so400m-patch14-384      & mistralai/Mistral-7B-v0.1                                           & 8B           \\
                                  & \multicolumn{3}{l}{\url{https://huggingface.co/HuggingFaceM4/idefics2-8b-base}}                                                  \\\midrule
                                  & EVA-CLIP                              & LLaMA                                                               & 14B          \\
\multirow{-2}{*}{Emu1}            & \multicolumn{3}{l}{\url{https://huggingface.co/BAAI/Emu/blob/main/Emu-pretrain.pt}} \\\bottomrule                                         
\end{tabular}%
}
\caption{Information of tested multimodal LLMs, their visual encoder, text models, number of parameters and the download links on Hugging face.}
\label{tab:model_details}
\end{table*}

\begin{figure*}[t!]
     \centering
     \includegraphics[width=\linewidth]{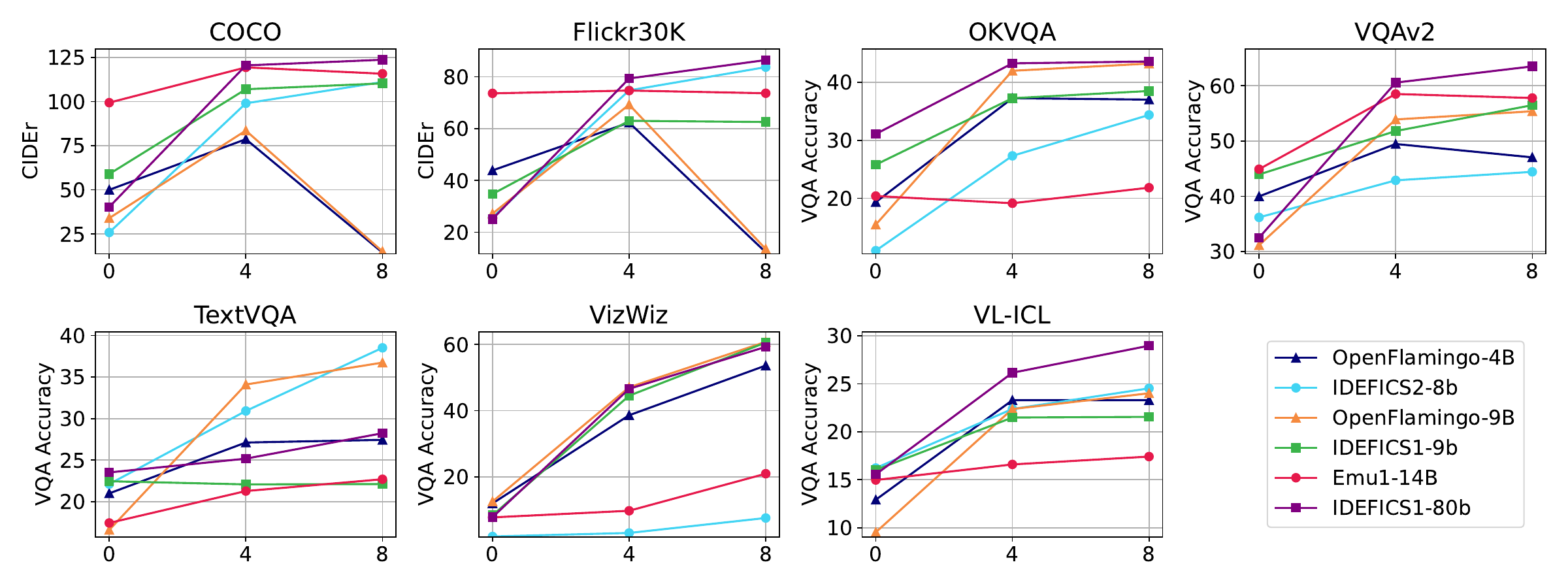}
        \caption{Evaluation on existing ICL tasks including image captioning (COCO and Flickr30K), general-purpose VQA (OKVQA, VQAv2, TextVQA and VizWiz) and recently released VL-ICL benchmark. We observe ICL abilities in general, with different levels across tasks, models and demonstration amounts: Emu1 benefits less from provided demonstrations compared with others, while two OpenFlamingo models exhibit worse performance when 8 demonstrations are provided. Refer to~\Cref{fig:prior_icl_gpt} for GPT-4o performance and ~\Cref{fig:vl_icl} for detailed results on each subset of the VL-ICL benchmark.}
        \label{fig:prior_icl}
\end{figure*}

\begin{figure}[]
     \centering
     \includegraphics[width=\linewidth]{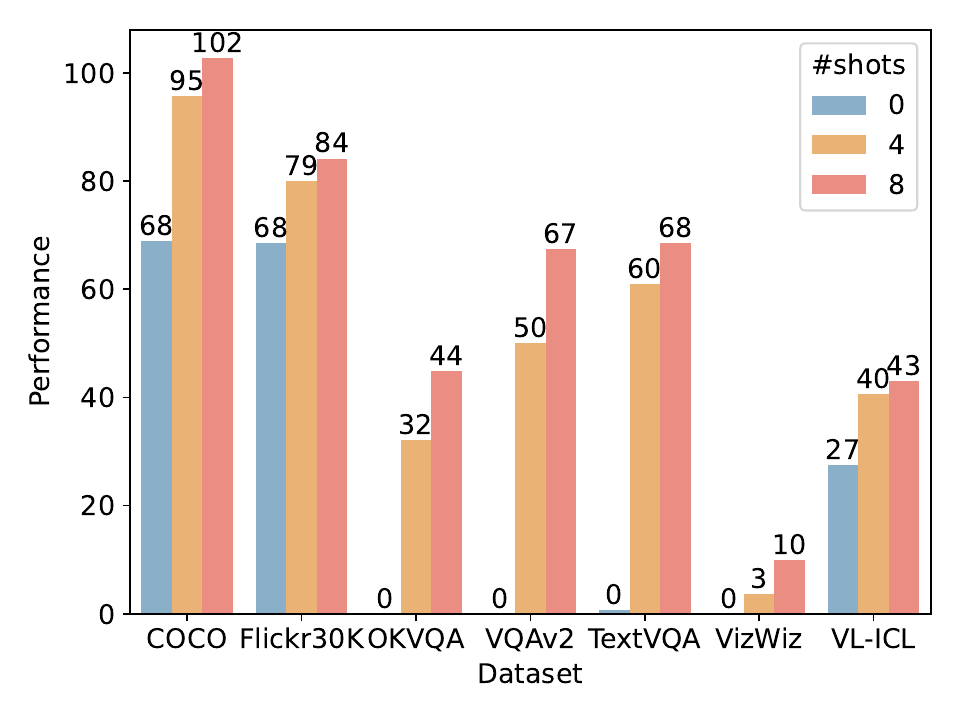}
        \caption{Performance of GPT-4o on existing ICL benchmarks. GPT-4o obtains much better performance when more demonstrations are given as the context.\protect\footnotemark}
        \label{fig:prior_icl_gpt}
\end{figure}

\begin{figure*}[t!]
     \centering
    \begin{subfigure}[b]{\textwidth}
         \centering
         \includegraphics[width=\linewidth]{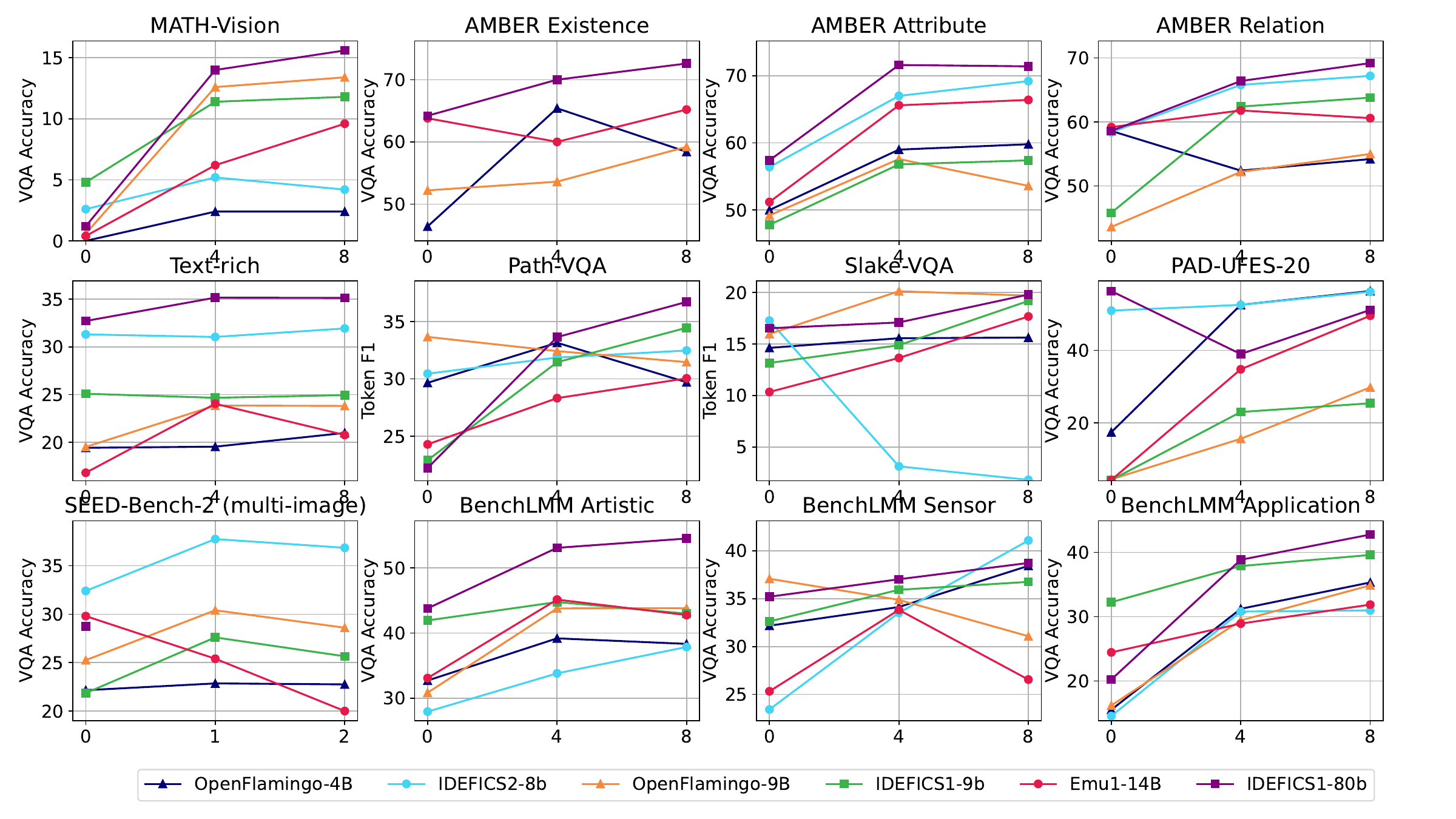}\caption{Pretrained Models}
         \label{fig:more_nogpt}
     \end{subfigure}
         \begin{subfigure}[b]{\textwidth}
         \centering
         \includegraphics[width=\linewidth]{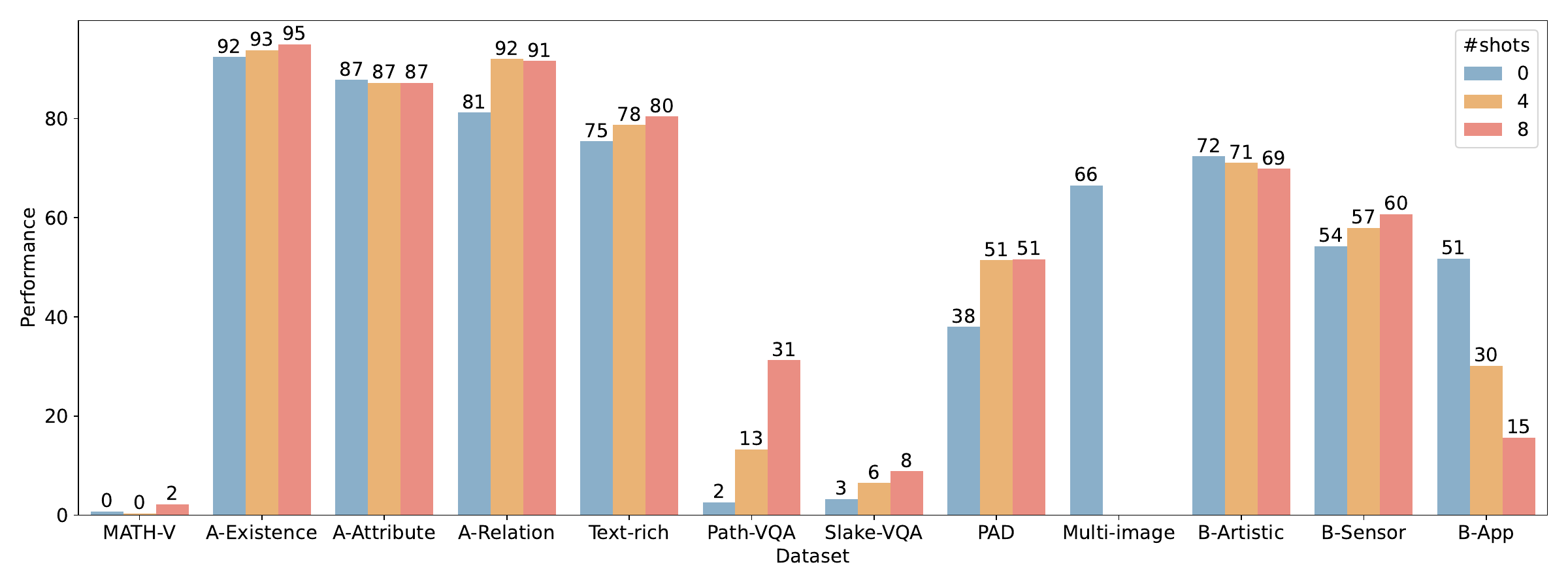}\caption{GPT-4o}
         \label{fig:more_gpt}
     \end{subfigure}
        \caption{More comprehensive ICL capability evaluation of pretrained models (top 3 row) and GPT-4o (bottom row) on recently proposed benchmarks. Pretrained models exhibit ICL abilities across different tasks, while GPT-4o achieves much higher zero-shot performance but benefits merely from provided demonstrations.}
        \label{fig:new_icl}
\end{figure*}

\begin{figure*}[t!]
     \centering
     \includegraphics[width=\linewidth]{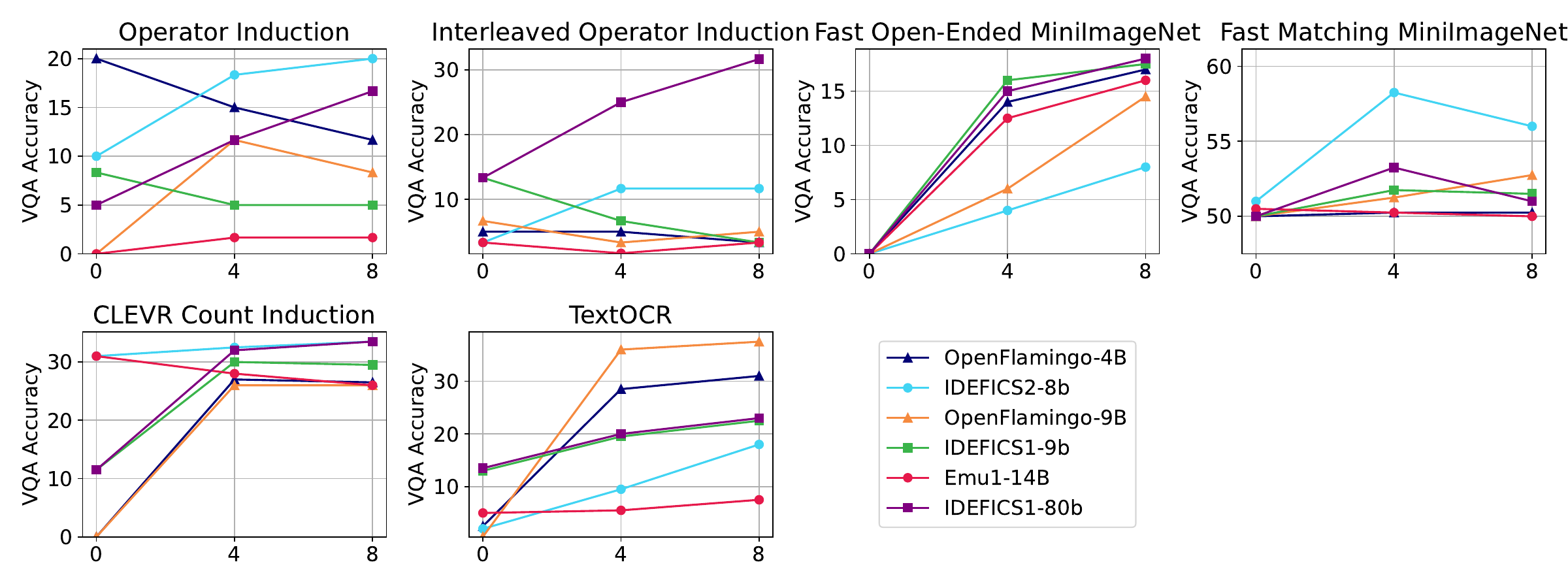}
        \caption{Detailed evaluation on VL-ICL Bench for multimodal in-context learning.}
        \label{fig:vl_icl}
\end{figure*}
\begin{figure}[t!]
     \centering
     \includegraphics[width=\linewidth]{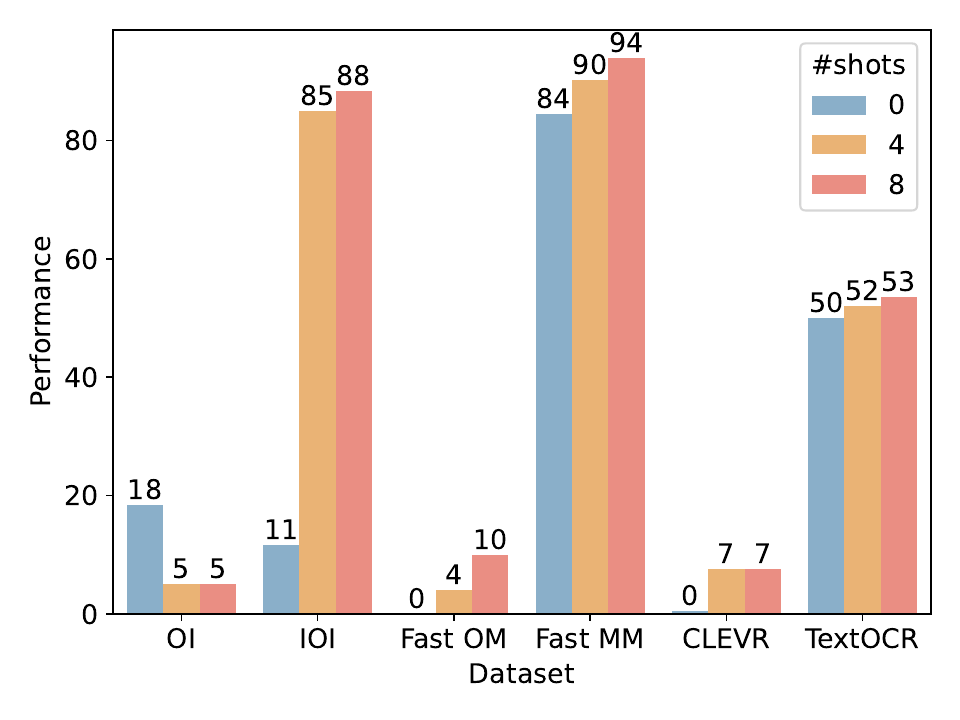}
        \caption{Detailed performance of GPT-4o on VL-ICL Bench for multimodal in-context learning.}
        \label{fig:vl_icl_gpt}
\end{figure}

\begin{figure*}[t!]
     \centering
         \begin{subfigure}[b]{\textwidth}
         \centering
         \includegraphics[width=\linewidth]{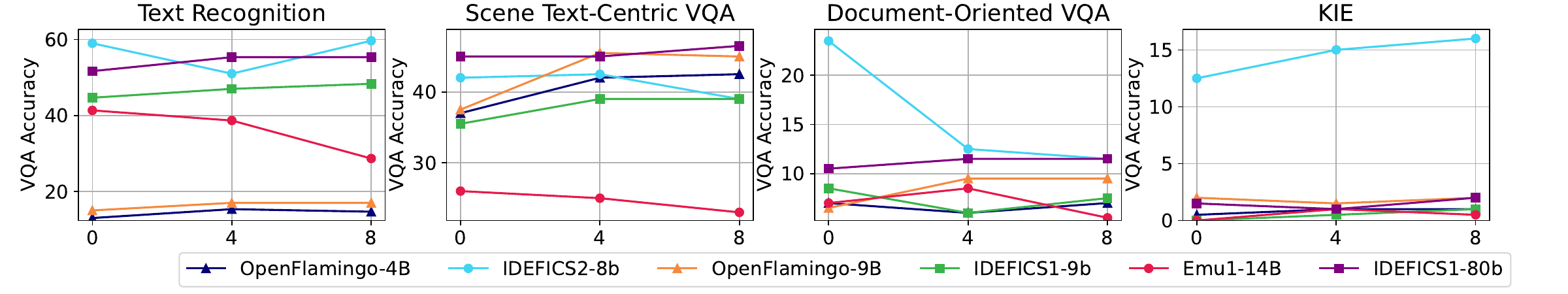}
         \label{fig:ocrbench}
     \end{subfigure}
         \begin{subfigure}[b]{\textwidth}
         \centering
         \includegraphics[width=\linewidth]{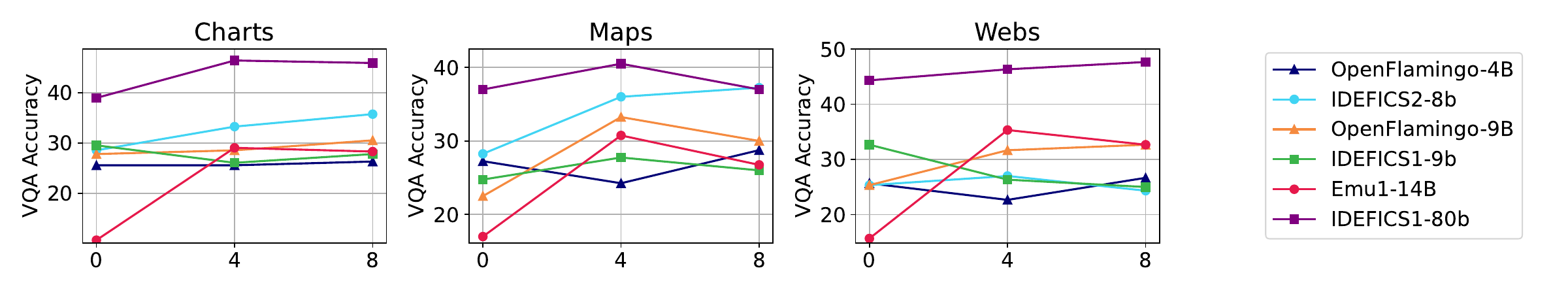}
         \label{fig:seedbench2plus}
     \end{subfigure}
        \caption{Detailed evaluation on OCRBench (top) and  SEED-Bench-2-Plus (bottom) for accessing Optical Character Recognition (OCR) capabilities.}
        \label{fig:ocr}
\end{figure*}

\begin{figure*}[t!]
     \centering
     \includegraphics[width=\linewidth]{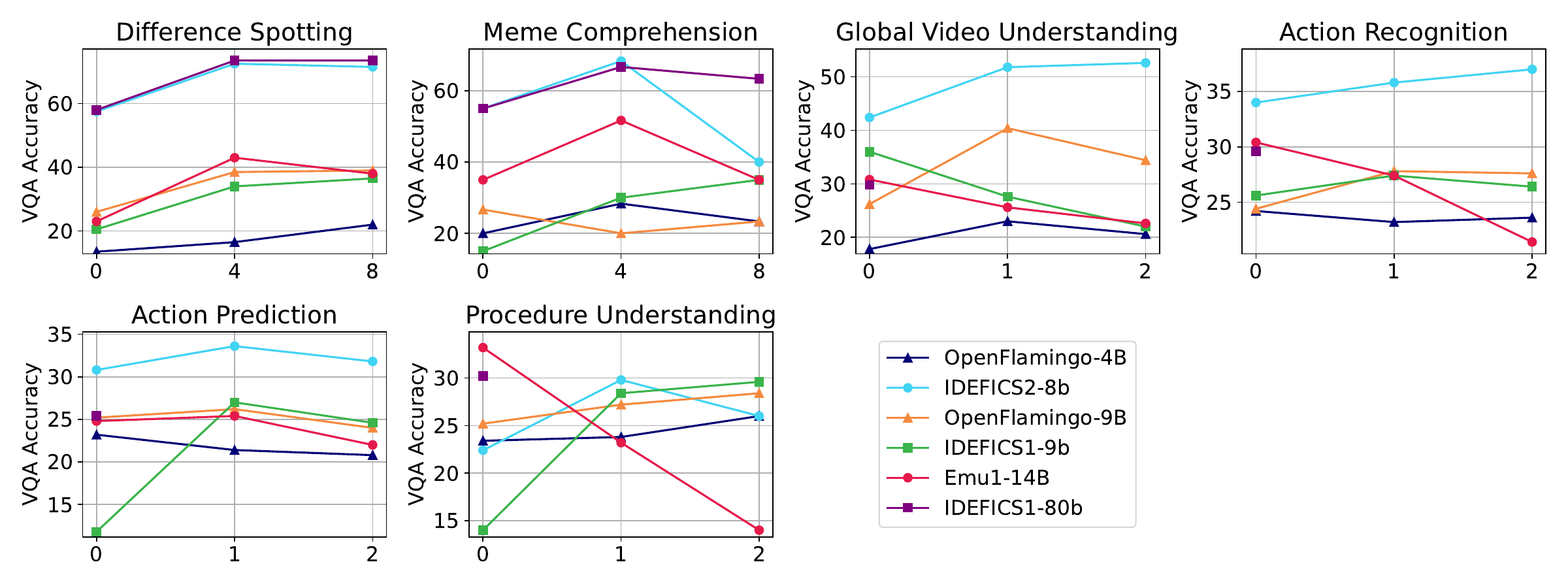}
        \caption{Detailed evaluation on SEED-Bench-2 for the ability to comprehend multiple images and texts.}
        \label{fig:seedbench2_nogpt}
\end{figure*}

\begin{figure*}[t!]
     \centering
     \includegraphics[width=\linewidth]{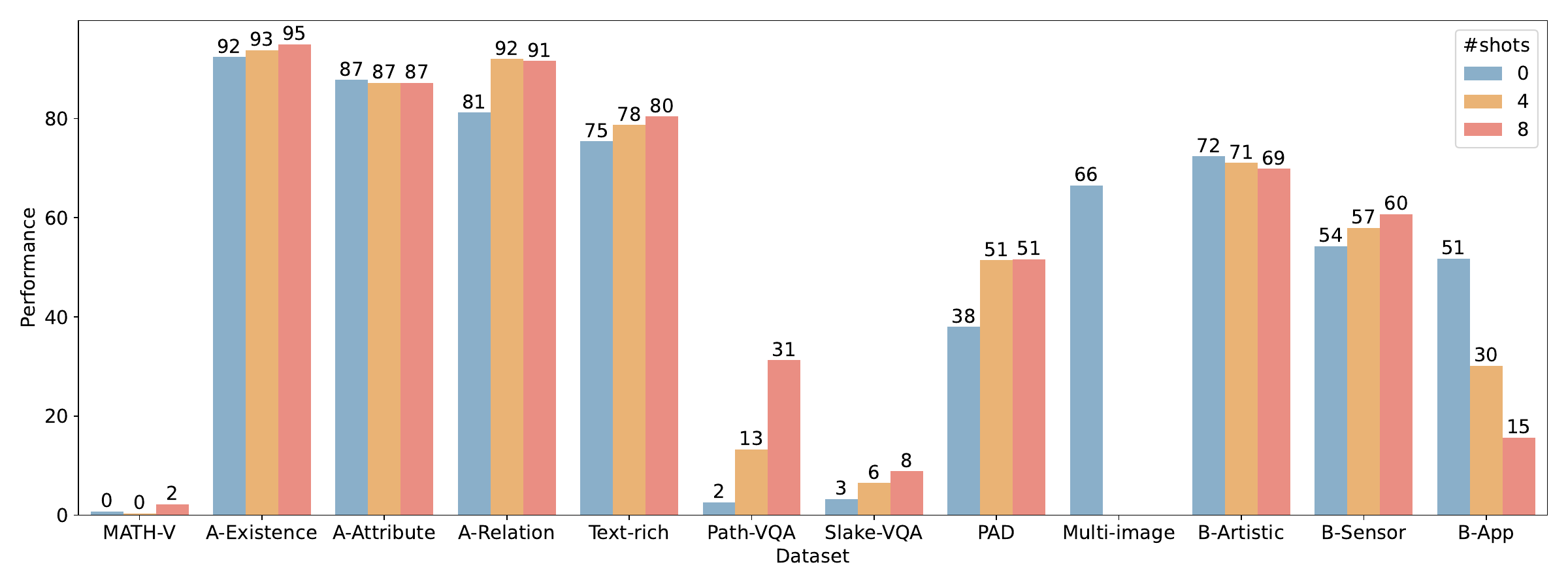}
        \caption{Detailed evaluation of GPT-4o on OCRBench (top),  SEED-Bench-2-Plus (middle) and SEED-Bench-2 (bottom).}
        \label{fig:seedbench2_gpt}
\end{figure*}
% ablation and demo selection
%1.openflamingo-4b
\begin{figure*}[t!]
     \centering
    \begin{subfigure}[b]{\textwidth}
         \centering
         \includegraphics[width=\linewidth]{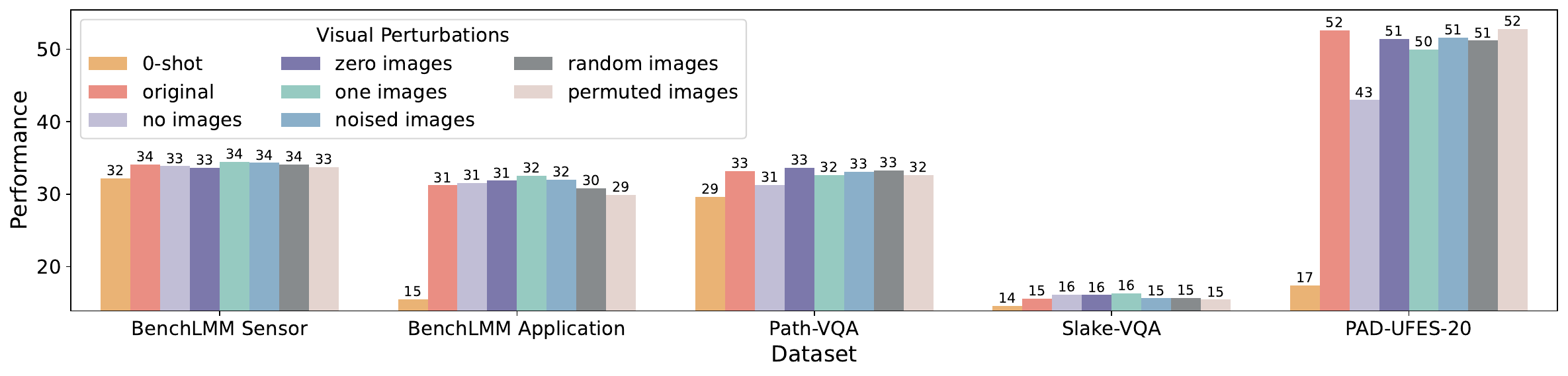}\caption{Visual Modality}
         \label{fig:visual_openflamingo2_4B}
     \end{subfigure}
         \begin{subfigure}[b]{\textwidth}
         \centering
         \includegraphics[width=\linewidth]{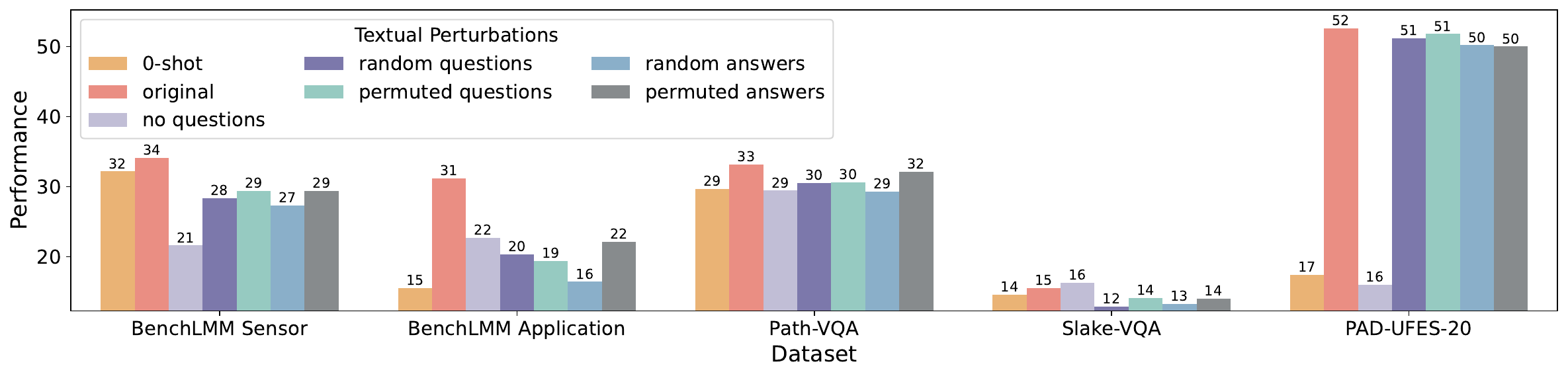}\caption{Textual Modality}
         \label{fig:textual_openflamingo2_4B}
     \end{subfigure}
              \begin{subfigure}[b]{\textwidth}
         \centering
         \includegraphics[width=\linewidth]{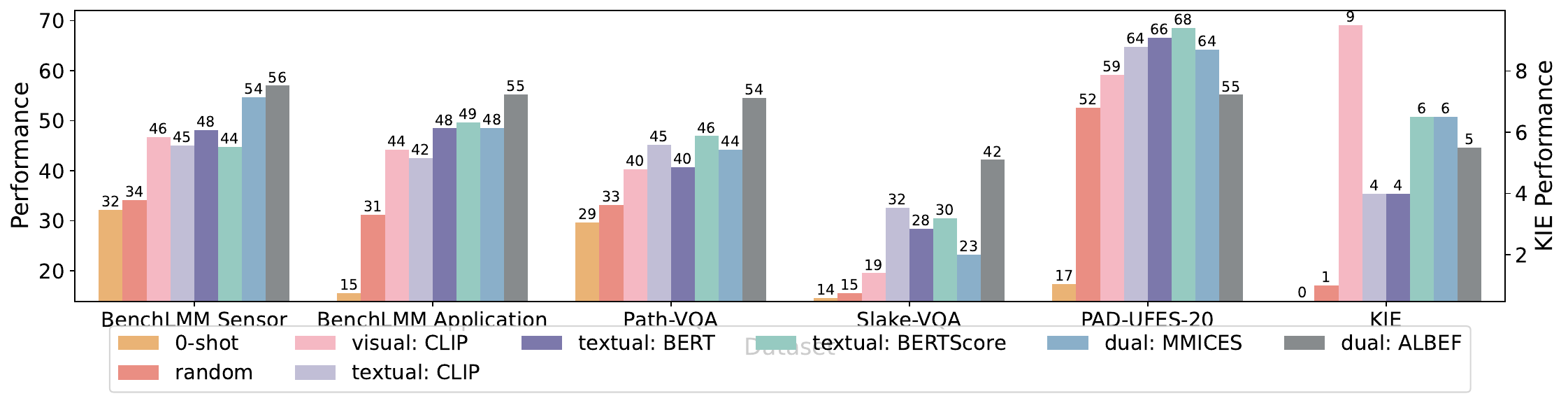}\caption{Textual Modality}
         \label{fig:selection_openflamingo2_4B}
     \end{subfigure}
        \caption{ICL performance of \textbf{OpenFlamingo2-4B} against perturbations over visual (top) and textual (middle) modalities and different demonstration selection strategies (bottom).}
        \label{fig:ablation_selection_openflamingo2_4B}
\end{figure*}
%2.idefics2-8b-base
\begin{figure*}[t!]
     \centering
    \begin{subfigure}[b]{\textwidth}
         \centering
         \includegraphics[width=\linewidth]{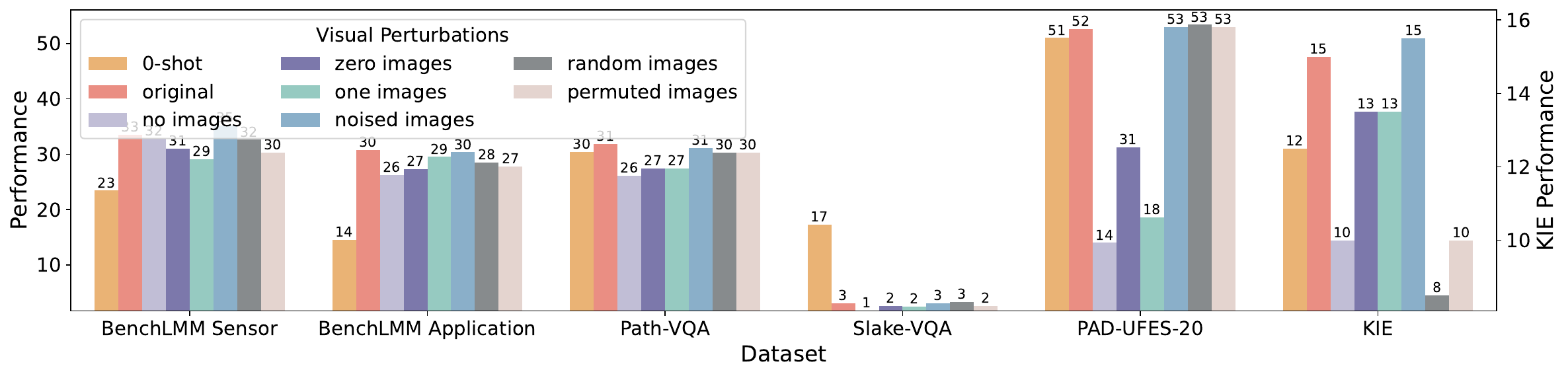}\caption{Visual Modality}
         \label{fig:visual_idefics2-8b-base}
     \end{subfigure}
         \begin{subfigure}[b]{\textwidth}
         \centering
         \includegraphics[width=\linewidth]{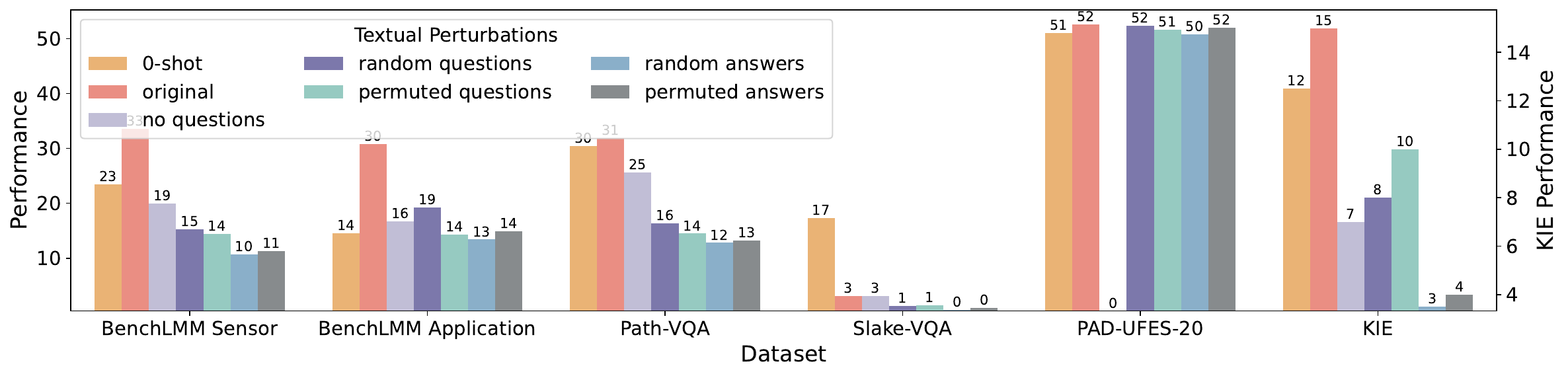}\caption{Textual Modality}
         \label{fig:textual_idefics2-8b-base}
     \end{subfigure}
              \begin{subfigure}[b]{\textwidth}
         \centering
         \includegraphics[width=\linewidth]{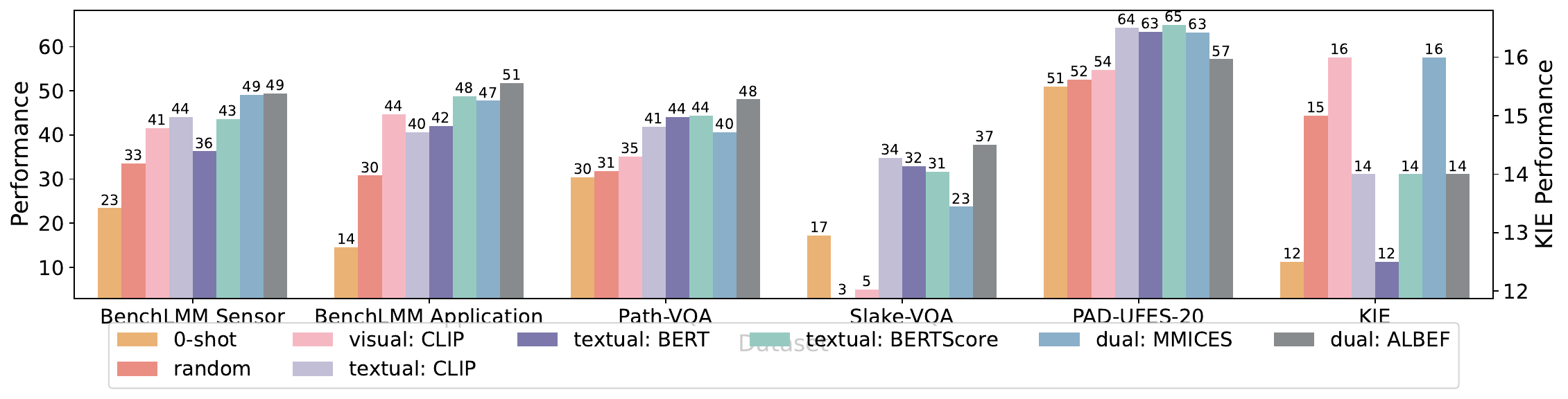}\caption{Textual Modality}
         \label{fig:selection_idefics2-8b-base}
     \end{subfigure}
        \caption{ICL performance of \textbf{IDEFICS2-8b} against perturbations over visual (top) and textual (middle) modalities and different demonstration selection strategies (bottom).}
        \label{fig:ablation_selection_idefics2-8b-base}
\end{figure*}
%3.openflamingo2
\begin{figure*}[t!]
     \centering
    \begin{subfigure}[b]{\textwidth}
         \centering
         \includegraphics[width=\linewidth]{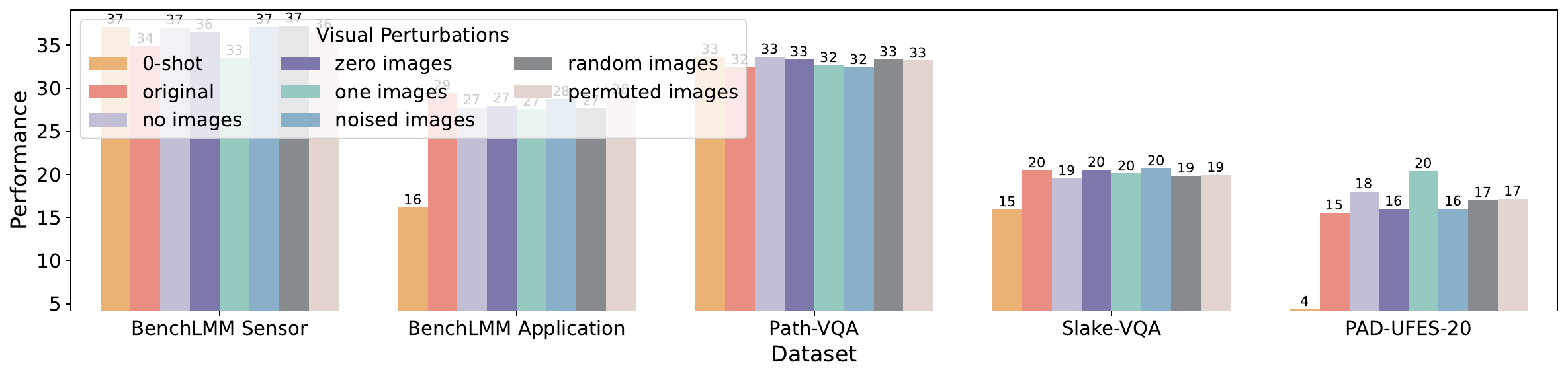}\caption{Visual Modality}
         \label{fig:visual_openflamingo2}
     \end{subfigure}
         \begin{subfigure}[b]{\textwidth}
         \centering
         \includegraphics[width=\linewidth]{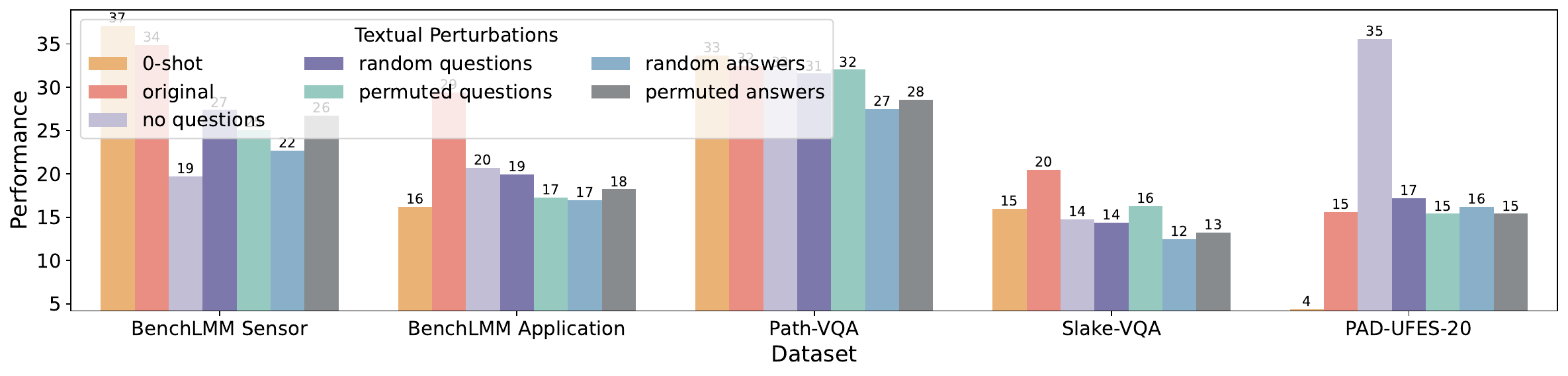}\caption{Textual Modality}
         \label{fig:textual_openflamingo2}
     \end{subfigure}
              \begin{subfigure}[b]{\textwidth}
         \centering
         \includegraphics[width=\linewidth]{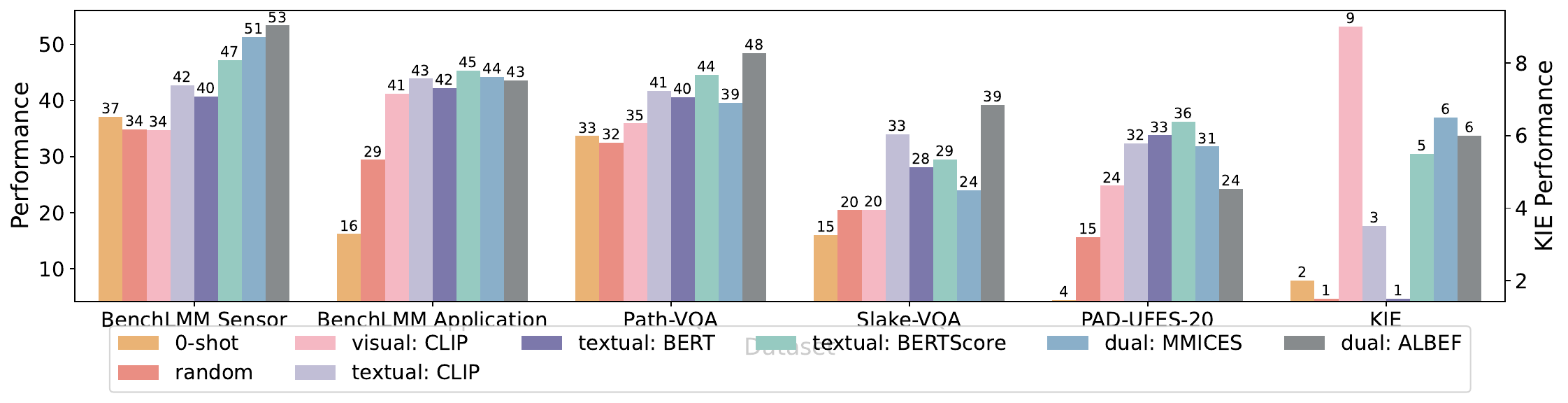}\caption{Textual Modality}
         \label{fig:selection_openflamingo2}
     \end{subfigure}
        \caption{ICL performance of \textbf{OpenFlamingo-9B} against perturbations over visual (top) and textual (middle) modalities and different demonstration selection strategies (bottom).}
        \label{fig:ablation_selection_openflamingo2}
\end{figure*}
%4.idefics-9b
\begin{figure*}[t!]
     \centering
    \begin{subfigure}[b]{\textwidth}
         \centering
         \includegraphics[width=\linewidth]{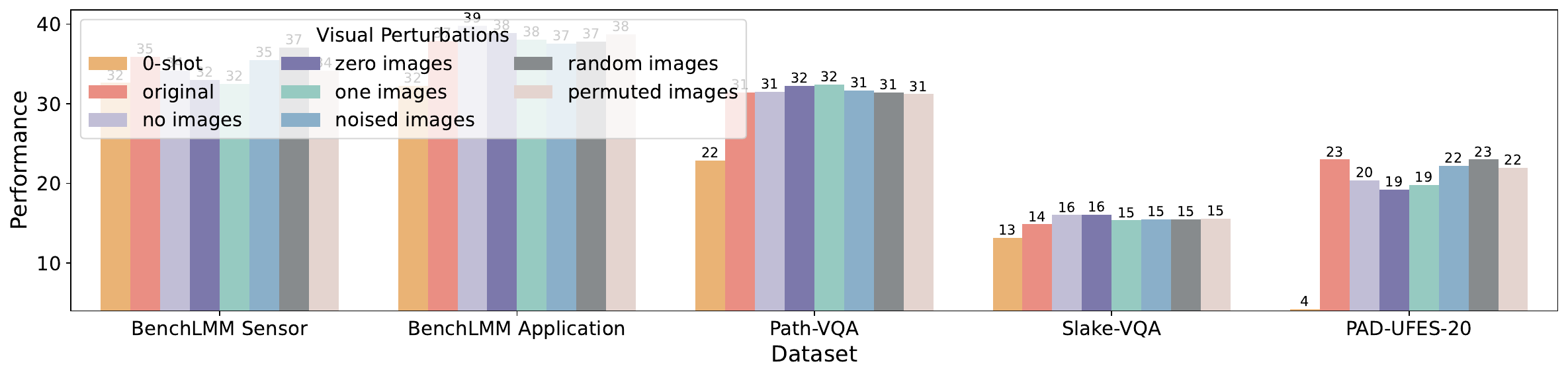}\caption{Visual Modality}
         \label{fig:visual_idefics-9b}
     \end{subfigure}
         \begin{subfigure}[b]{\textwidth}
         \centering
         \includegraphics[width=\linewidth]{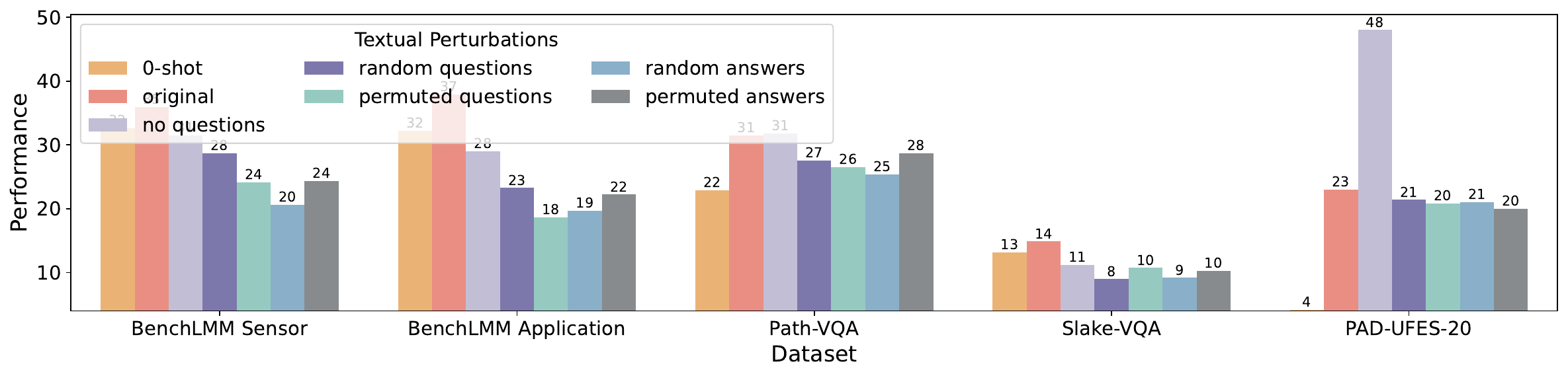}\caption{Textual Modality}
         \label{fig:textual_idefics-9b}
     \end{subfigure}
              \begin{subfigure}[b]{\textwidth}
         \centering
         \includegraphics[width=\linewidth]{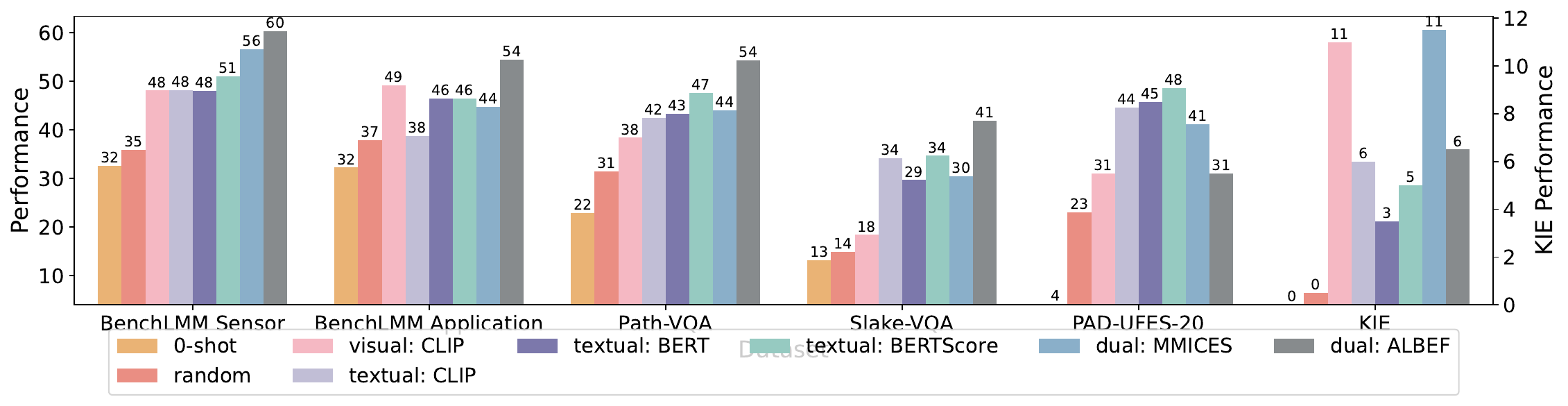}\caption{Textual Modality}
         \label{fig:selection_idefics-9b}
     \end{subfigure}
        \caption{ICL performance of \textbf{IDEFICS1-9b} against perturbations over visual (top) and textual (middle) modalities and different demonstration selection strategies (bottom).}
        \label{fig:ablation_selection_idefics-9b}
\end{figure*}
%5.Emu1
\begin{figure*}[t!]
     \centering
    \begin{subfigure}[b]{\textwidth}
         \centering
         \includegraphics[width=\linewidth]{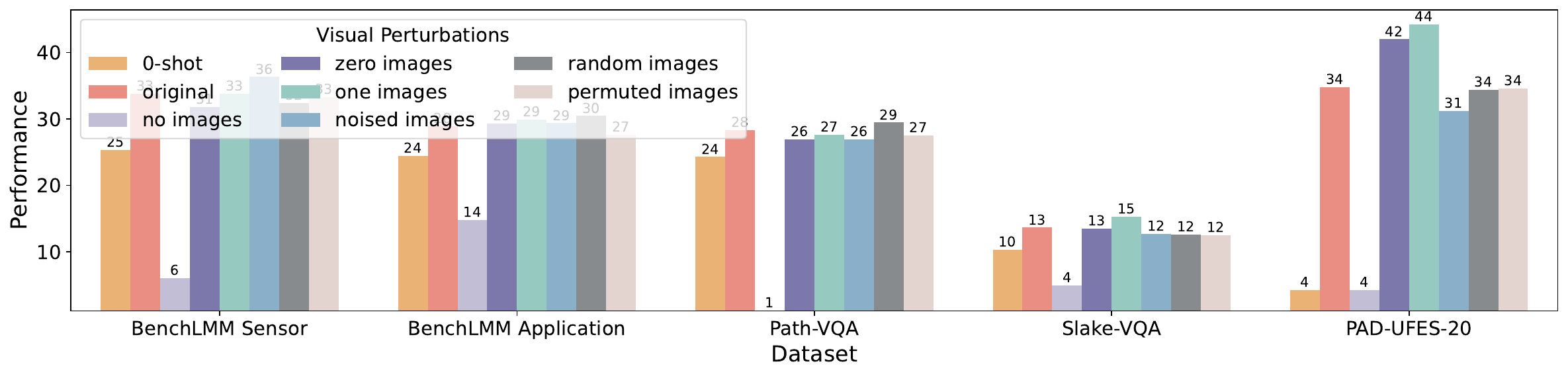}\caption{Visual Modality}
         \label{fig:visual_Emu1}
     \end{subfigure}
         \begin{subfigure}[b]{\textwidth}
         \centering
         \includegraphics[width=\linewidth]{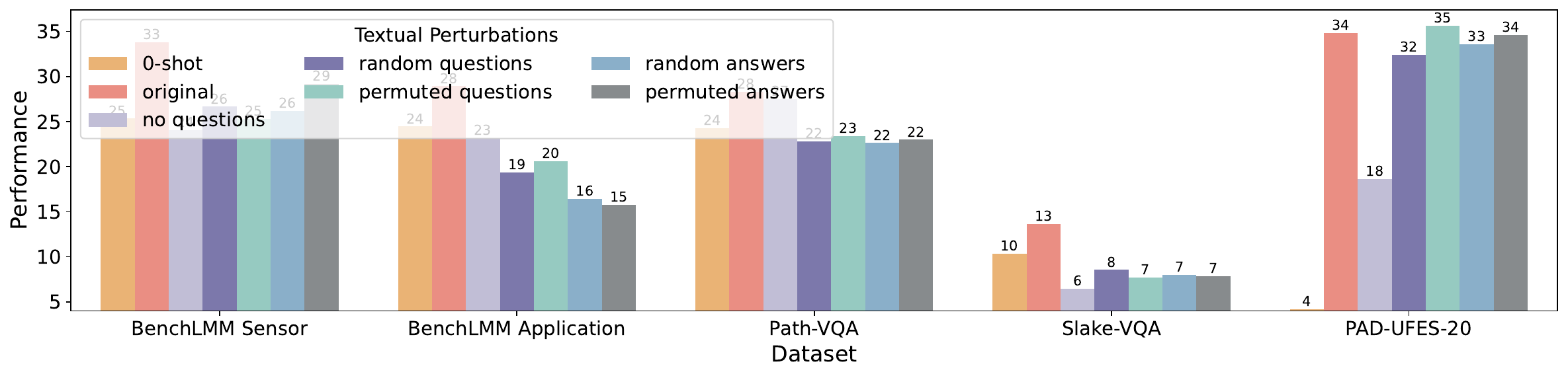}\caption{Textual Modality}
         \label{fig:textual_Emu1}
     \end{subfigure}
              \begin{subfigure}[b]{\textwidth}
         \centering
         \includegraphics[width=\linewidth]{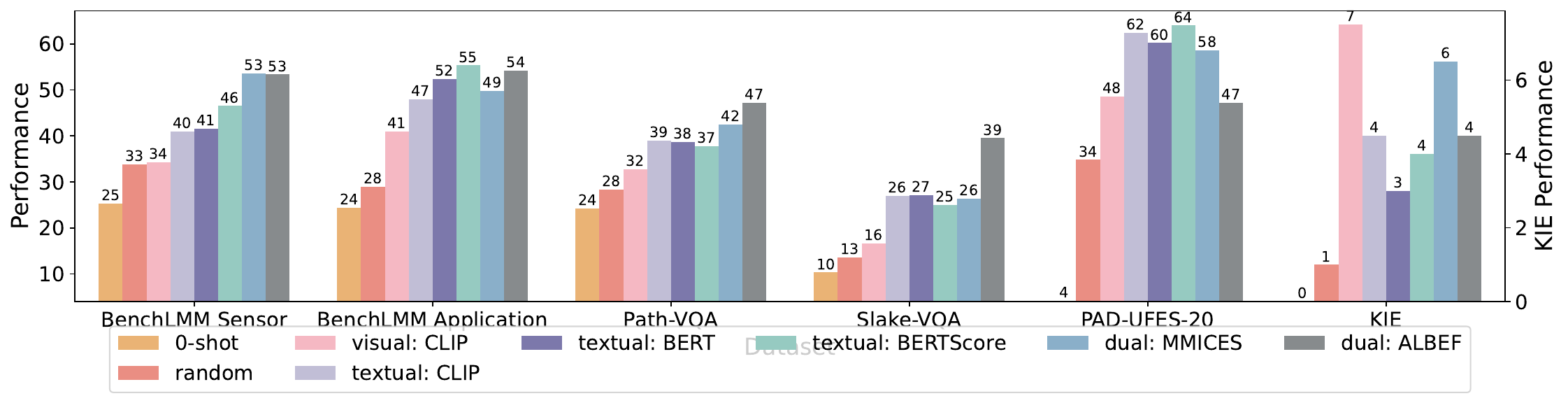}\caption{Textual Modality}
         \label{fig:selection_Emu1}
     \end{subfigure}
        \caption{ICL performance of \textbf{Emu1-14B} against perturbations over visual (top) and textual (middle) modalities and different demonstration selection strategies (bottom).}
        \label{fig:ablation_selection_Emu1}
\end{figure*}
% flip experiments
\begin{figure*}[t!]
     \centering
              \begin{subfigure}[b]{\textwidth}
         \centering
         \includegraphics[width=\linewidth]{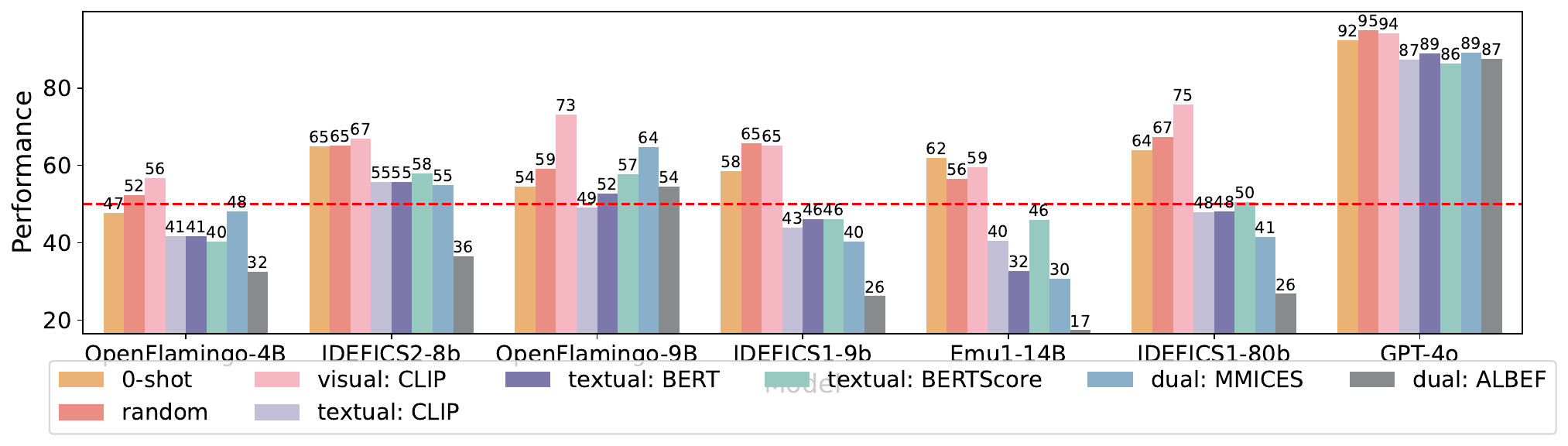}\caption{AMBER Existence}
         \label{fig:AMBER_existence_mix_flip}
     \end{subfigure}
              \begin{subfigure}[b]{\textwidth}
         \centering
         \includegraphics[width=\linewidth]{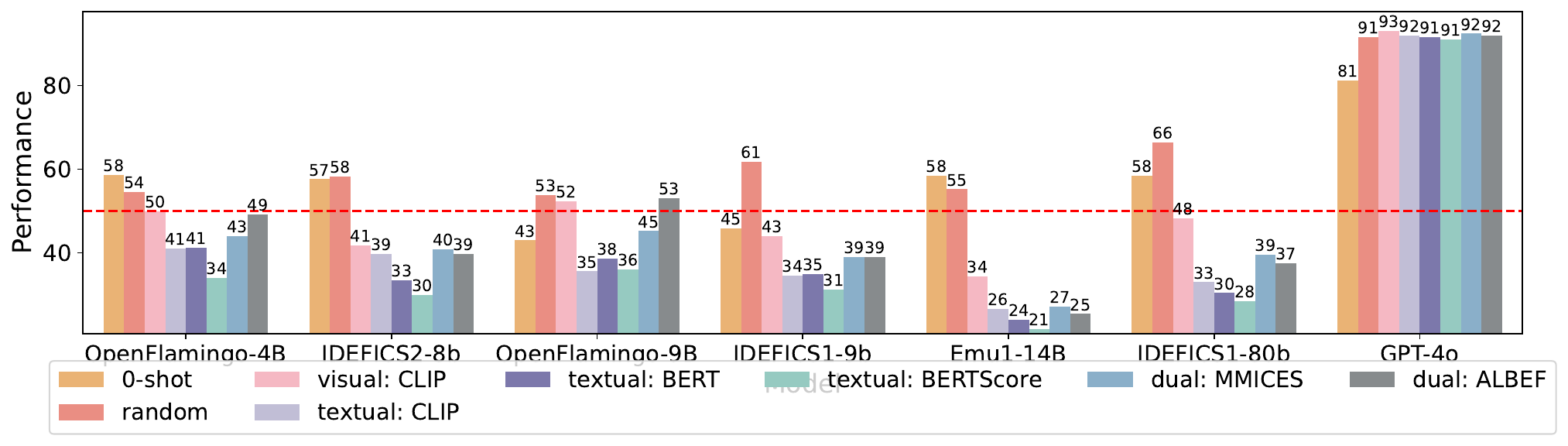}\caption{AMBER Relation}
         \label{fig:AMBER_relation_flip}
     \end{subfigure}
        \caption{The abilities to capture inductive biases with flipped in-context annotations on AMBER Existence and Relation dataset. }
        \label{fig:AMBER_flip}
\end{figure*}

\end{document}